\documentclass[a4paper,fleqn]{cas-dc}

\usepackage[numbers,sort]{natbib}
\usepackage{amssymb}
\usepackage{multirow}
\usepackage[normalem]{ulem}
\useunder{\uline}{\ul}{}
\usepackage{subfigure}
\usepackage{color}
\usepackage{lineno,hyperref}
\usepackage{booktabs}
\usepackage{amsmath}
\usepackage{lscape}
\usepackage{array}
\usepackage{float}
\usepackage{graphicx}
\usepackage{balance}

\def\tsc#1{\csdef{#1}{\textsc{\lowercase{#1}}\xspace}}
\tsc{WGM}
\tsc{QE}

\begin{document}
\let\WriteBookmarks\relax
\def\floatpagepagefraction{1}
\def\textpagefraction{.001}
\let\printorcid\relax

\shorttitle{Is ChatGPT a Good Personality Recognizer? A Preliminary Study}

\shortauthors{Yu Ji et~al.}

\title [mode = title]{Is ChatGPT a Good Personality Recognizer? A Preliminary Study}  

\author[1,2]{Yu Ji}[orcid=0000-0001-6048-9184]
\ead{52205901009@stu.ecnu.edu.cn}

\author[2,3]{Wen Wu}[orcid=0000-0002-2132-5993]
\cormark[1]
\ead{wwu@cc.ecnu.edu.cn}
\cortext[1]{Corresponding author}

\author[4]{Hong Zheng}

\author[3]{Yi Hu}

\author[3]{Xi Chen}

\author[1,2]{Liang He}

\affiliation[1]{organization={Institute of AI Education, East China Normal University},
             city={Shanghai},
             country={China}}

\affiliation[2]{organization={School of Computer Science and Technology, East China Normal University},
             city={Shanghai},
             country={China}}

\affiliation[3]{organization={Shanghai Key Laboratory of Mental Health and Psychological Crisis Intervention, School of Psychology and Cognitive Science, East China Normal University},
             city={Shanghai},
             country={China}}

\affiliation[4]{organization={Shanghai Changning Mental Health Center},
             city={Shanghai},
             country={China}}

\begin{abstract}
In recent years, personality has been regarded as a valuable personal factor being incorporated into numerous tasks such as sentiment analysis and product recommendation. This has led to widespread attention to text-based personality recognition task, which aims to identify an individual's personality based on given text. Considering that ChatGPT has recently exhibited remarkable abilities on various natural language processing tasks, we provide a preliminary evaluation of ChatGPT on text-based personality recognition task for generating effective personality data. Concretely, we employ a variety of prompting strategies to explore ChatGPT's ability in recognizing personality from given text, especially the level-oriented prompting strategy we designed for guiding ChatGPT in analyzing given text at a specified level. The experimental results on two representative real-world datasets reveal that ChatGPT with zero-shot chain-of-thought prompting exhibits impressive personality recognition ability and is capable to provide natural language explanations through text-based logical reasoning. Furthermore, by employing the level-oriented prompting strategy to optimize zero-shot chain-of-thought prompting, the performance gap between ChatGPT and corresponding state-of-the-art model has been narrowed even more. However, we observe that ChatGPT shows unfairness towards certain sensitive demographic attributes such as gender and age. Additionally, we discover that eliciting the personality recognition ability of ChatGPT helps improve its performance on personality-related downstream tasks such as sentiment classification and stress prediction.
\end{abstract}



\maketitle

\section{Introduction}
As one of the basic individual characteristics, personality describes the relatively stable pattern of individual w.r.t. her/his behavior, thought, and emotion \cite{diener2019personality}. In recent years, an increasing number of researchers have considered personality as a valuable factor and incorporated it into various tasks (e.g., machine translation \cite{mirkin2015motivating,rabinovich2017personalized}, product recommendation \cite{liu2023emotion,martijn2022knowing}, sentiment analysis \cite{lin2017personality}, and mental health analysis \cite{jaiswal2019automatic}), resulting in significant performance improvements. In order to automatically obtain large-scale user personality, text-based personality recognition task is designed to infer user personality based on given user-generated text \cite{lin2023novel,DBLP:conf/ijcai/ZhuHGPW22,zhu2022lexical}. With the rapid developments of pre-trained Large Language Models (LLMs) (e.g., BERT \cite{kenton2019bert}, RoBERTa \cite{liu2019roberta}, GPT-3 \cite{brown2020language},PaLM \cite{narang2022pathways}, and LLaMA \cite{touvron2023llama}), more and more LLMs-based methods have been proposed for text-based personality detection task and have achieved remarkable performance improvements \cite{jain2022personality,jun2021personality}.

More recently, ChatGPT\footnote{https://chat.openai.com/} has attracted a considerable amount of attention with its impressive general language processing ability \cite{qin2023chatgpt}, sparking exploration into its capability boundaries \cite{sun2023chatgpt,wang2023chatgpt}. Several works have provided a preliminary evaluation of ChatGPT on various common tasks such as machine translation \cite{jiao2023chatgpt}, product recommendation \cite{liu2023chatgpt}, sentiment analysis \cite{wang2023chatgpt}, and mental health analysis \cite{yang2023evaluations}. Therefore, in this work, we are interested in evaluating the performance of ChatGPT on text-based personality recognition task for generating effective personality data. We also would like to see whether eliciting the personality recognition ability of ChatGPT contributes to improving its performance on other downstream tasks. Concretely, we raise the following Research Questions ({\bf RQs}):

{\bf RQ1}: How do different prompting strategies affect ChatGPT's ability to identify personality?

{\bf RQ2}: How unfair is ChatGPT when serving as a personality recognizer on various sensitive demographic attributes?

{\bf RQ3}: Does the personality inferred by ChatGPT help improve its performance on other downstream tasks?

To answer these research questions, we conduct experiments on two representative text-based personality recognition datasets (i.e., Essays and PAN) to compare the performance of ChatGPT, traditional neural network (e.g., Recurrent Neural Network (RNN)), fine-tuned RoBERTa, and corresponding State-Of-The-Art (SOTA) model. Specifically, we adopt three classic prompting strategies to elicit the personality recognition ability of ChatGPT, including zero-shot prompting, zero-shot Chain-of-Thought (CoT) prompting, and one-shot prompting. Furthermore, considering that researchers typically analyze texts at different levels (e.g., word level, sentence level, and document level) to obtain valuable text information \cite{chen2016neural,le2014distributed,ezaldeen2022hybrid,ristoski2020large}, we design zero-shot level-oriented CoT prompting to guide ChatGPT in analyzing given text at a specified level, thereby gaining a more targeted understanding of given text and recognizing personality more precisely. According to the experimental results, our findings can be summarized as follows:

(1) Among the three classic prompting strategies, zero-shot CoT prompting can better elicit ChatGPT's ability to predict personality based on given text, resulting in its optimal overall performance on the two datasets, although there is still a certain gap in performance compared to the SOTA model. Additionally, ChatGPT with zero-shot CoT prompting could generate more natural language explanations by text-based logical reasoning, enhancing the interpretability of the prediction results. Furthermore, with the assistance of zero-shot level-oriented CoT prompting, ChatGPT could perform more targeted text analysis, enabling it to complete more accurate personality prediction.

(2) ChatGPT exhibits unfairness to some sensitive demographic attributes on text-based personality recognition task. Based on ChatGPT's analysis, the woman group is more likely to have high levels of Openness, Conscientiousness, and Agreeableness when compared to the man group. Besides, relative to the younger group, the elderly group has a higher likelihood to have low Openness.

(3) The personality inferred by ChatGPT could enhance its performance on sentiment classification task and stress prediction task, which may provide new insights for other personality-related tasks (e.g., machine translation and product recommendation).

In the following sections, we first introduce related work regarding personality recognition in Section~\ref{sec-related-work}. After that, we present the details of our experimental design and analyze the experimental results in Section~\ref{sec-experiments}. Finally, we conclude the paper and indicate some future directions in Section~\ref{sec-conclusion}.

\section{Background and Related Work}\label{sec-related-work}

Big-Five Factor (BFF) model and Myers-Briggs Type Indicator (MBTI) are two most popular personality assessment models \cite{celli2018big}. To be specific, BFF model describes personality based on five traits: Openness (O), Conscientiousness (C), Extraversion (E), Agreeableness (A), and Neuroticism (N) \cite{digman1990personality}. Table~\ref{table-bff} shows the propensities of individuals under different personality traits and levels. On the contrary, MBTI describes personality according to four dimensions, including Extraversion/Introversion, Sensing/Intuition, Thinking/Feeling, and Judging/Perceiving \cite{pittenger1993measuring}. Compared to BFF model, MBTI still faces controversy within the academic community \cite{boyle1995myers,pittenger2005cautionary}. Hence, we adopt BFF model to describe individuals' personalities in this paper.

In recent years, an increasing number of researchers regarded Big-Five personality as a valuable personal factor and incorporated it into their models, resulting in significant performance improvements on various tasks \cite{ban2022knowledge,wu2018personalizing,klec2023beyond,xu2022recommendation}. For example, Wu et al. \cite{wu2018personalizing} adopted users' Big-Five personalities to personalize the recommendation diversity being tailored to the users' diversity needs. Ban et al. \cite{ban2022knowledge} utilized learners' Big-Five personalities to model the individual differences for better predicting the learners' knowledge levels. This has sparked researchers' interest in efficiently acquiring Big-Five personalities.

\begin{table}
\centering
\caption{Individual propensities under different personality traits and levels}\label{table-pro-per}
\begin{tabular}{ccl}
\toprule
\multicolumn{1}{l}{Personality Trait} & Level & Propensities           \\ \midrule
\multirow{2}{*}{O}              & High   & Creative, Open-minded    \\ 
                                & Low    & Reflective, Conventional \\  \midrule
\multirow{2}{*}{C}              & High   & Disciplined, Prudent     \\ 
                                & Low    & Careless, Impulsive      \\ \midrule
\multirow{2}{*}{E}              & High   & Sociable, Talkative      \\ 
                                & Low    & Reserved, Shy            \\ \midrule
\multirow{2}{*}{A}              & High   & Trusting, Cooperative       \\ 
                                & Low    &   Aggressive, Cold                       \\ \midrule
\multirow{2}{*}{N}              & High   & Worry, Sensitivity                         \\ 
                                & Low    &  Secure, Confident                        \\ \bottomrule
\end{tabular}
\label{table-bff}
\end{table}

The conventional approach to identify an individual's Big-Five personality is via personality questionnaires (e.g., NEO-FFI questionnaire \cite{costa1989neo}, BFI-44 \cite{john1991big}, BFI-10 \cite{rammstedt2007measuring}, and BFMS \cite{perugini2002analyzing}). These personality questionnaires are typically carefully designed by psychology experts and require individuals to rate their behaviors using Likert scales, which is time-consuming and labor-intensive \cite{han2020knowledge,tareaf2019facial}. In order to apply Big-Five personality on a large scale across various domains (e.g., machine translation \cite{mirkin2015motivating,rabinovich2017personalized}, product recommendation \cite{liu2023emotion,martijn2022knowing}, sentiment analysis \cite{lin2017personality}, and mental health analysis \cite{jaiswal2019automatic}), researchers attempted to implicitly obtain Big-Five personality from various User-Generated Content (UGC), including text \cite{lin2023novel,DBLP:conf/ijcai/ZhuHGPW22,zhu2022lexical,jain2022personality,jun2021personality}, handwriting \cite{gavrilescu2018predicting,ghali2022human,ji2023automatic}, speech \cite{perez2022automatic,sangeetha2020speech}, electroencephalography (EEG) \cite{bhardwaj2021eeg,li2022quantitative}, and so on. Due to substantial evidence from psychological research demonstrating the correlation between user-generated texts and users' Big-Five personalities \cite{park2015automatic,ireland2010language}, researchers made an extensive exploration of text-based personality recognition. However, the related methods normally regarded text-based personality recognition task as a special case of text classification. Most of them utilized machine learning algorithms to build personality recognizers with text features such as Linguistic Inquiry and Word Count (LIWC) \cite{pennebaker2015development,tandera2017personality} and Structured Programming for Linguistic Cue Extraction (SPLICE) \cite{moffitt2012structured,tadesse2018personality}. Furthermore, with the rapid development of deep learning, more and more methods using deep neural networks are proposed to solve text-based personality recognition task, as deep neural networks could extract high-order text features from user-generated text automatically \cite{majumder2017deep}. For example, Majumder et al. \cite{majumder2017deep} designed a deep convolutional neural network with Word2Vec embeddings \cite{mikolov2013efficient} for personality detection. Xue et al. \cite{xue2018deep} presented a two-level hierarchical neural network to learn the deep semantic representations of users' posts for recognizing users' Big-Five personalities. Lynn et al. \cite{lynn2020hierarchical} utilized message-level attention to learn the relative weight of users' posts for assessing users' Big-Five personalities. Zhu et al. \cite{DBLP:conf/ijcai/ZhuHGPW22} learned post embeddings by contrastive graph transformer network for personality detection. Zhu et al. \cite{zhu2022lexical} proposed a lexical psycholinguistic knowledge-guided graph neural network to enrich the semantics of users' posts with the personality lexicons. Recently, the remarkable performance enhancements achieved by LLMs in numerous Nature Language Processing (NLP) tasks \cite{sun2019fine,lin2022predictive,venugopalan2022enhanced} prompted researchers to explore the utilization of LLMs in text-based personality prediction task \cite{jain2022personality,jun2021personality}. For example, Mehta et al. \cite{mehta2020bottom} performed extensive experiments with BERT to arrive at the optimal configuration for personality detection. Ren et al. \cite{ren2021sentiment} leveraged BERT to generate sentence-level embedding for personality recognition, while a sentiment dictionary is used to consider sentiment information in the process of personality prediction.

Lately, the release of ChatGPT has drawn increasingly great attention due to the incredible general language processing ability of ChatGPT. Therefore, more and more researchers attempted to explore the capability boundaries of ChatGPT and evaluate it on various tasks, including machine translation \cite{jiao2023chatgpt}, product recommendation \cite{liu2023chatgpt}, sentiment analysis \cite{wang2023chatgpt}, mental health analysis \cite{yang2023evaluations}, and so on. Hence, in this work, we are interested in exploring the personality recognition ability of ChatGPT through different prompting strategies for obtaining effective personality data.

\begin{figure*}
	\centering
	\subfigure[Essays-Training]{\includegraphics[width=0.32\textwidth]{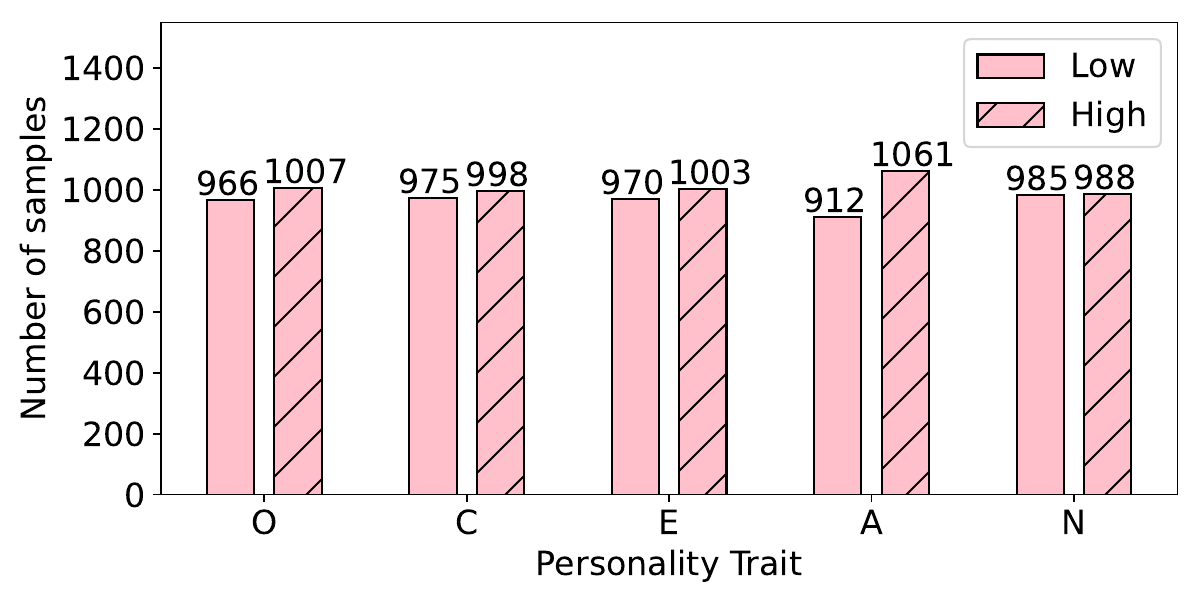}}
    \subfigure[Essays-Validation]{\includegraphics[width=0.32\textwidth]{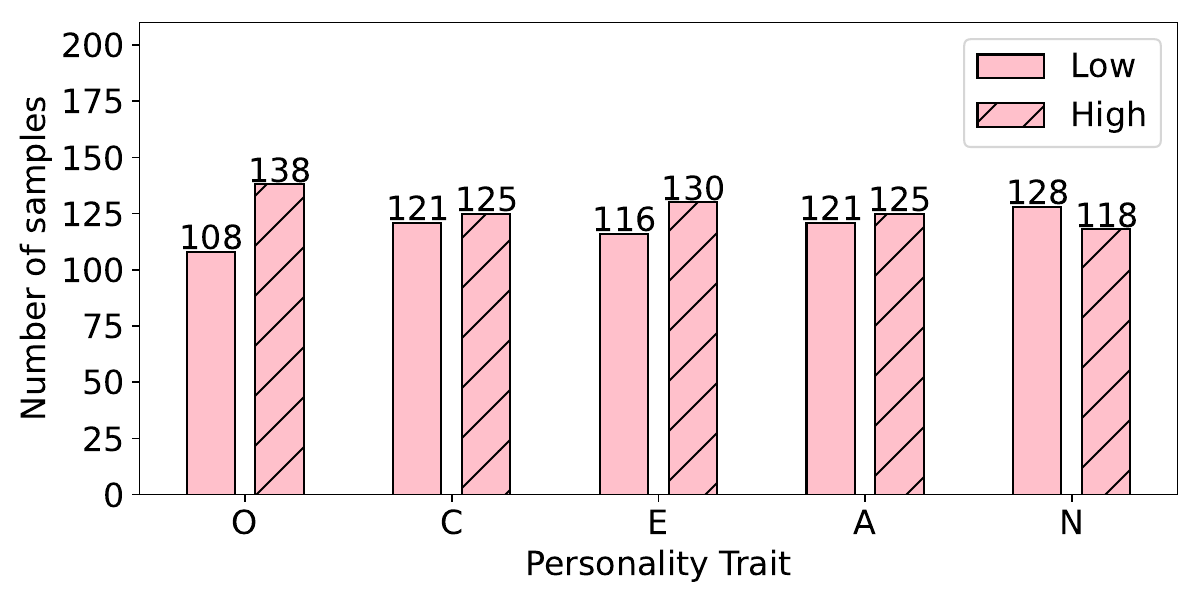}}
	\subfigure[Essays-Testing]{\includegraphics[width=0.32\textwidth]{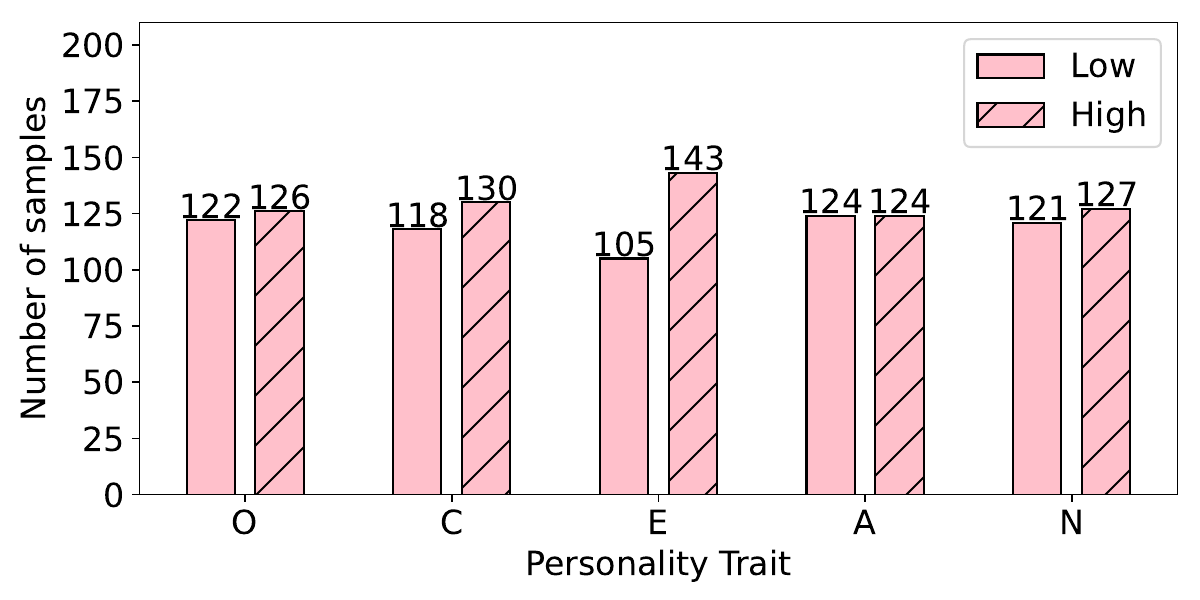}}

	\subfigure[PAN-Training]{\includegraphics[width=0.32\textwidth]{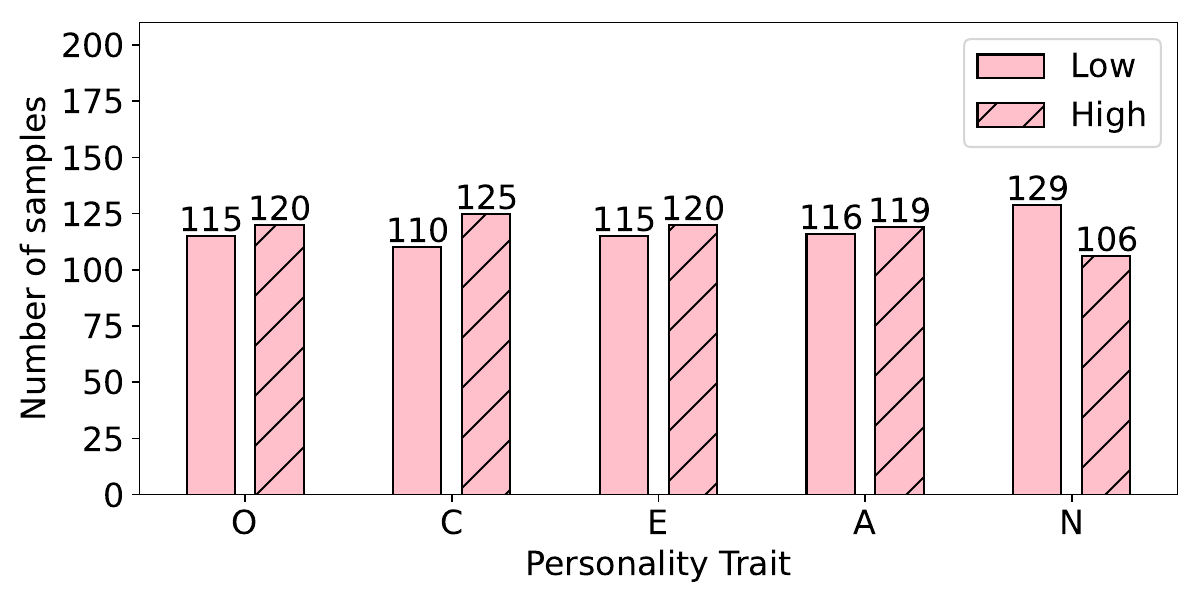}}
    \subfigure[PAN-Validation]{\includegraphics[width=0.32\textwidth]{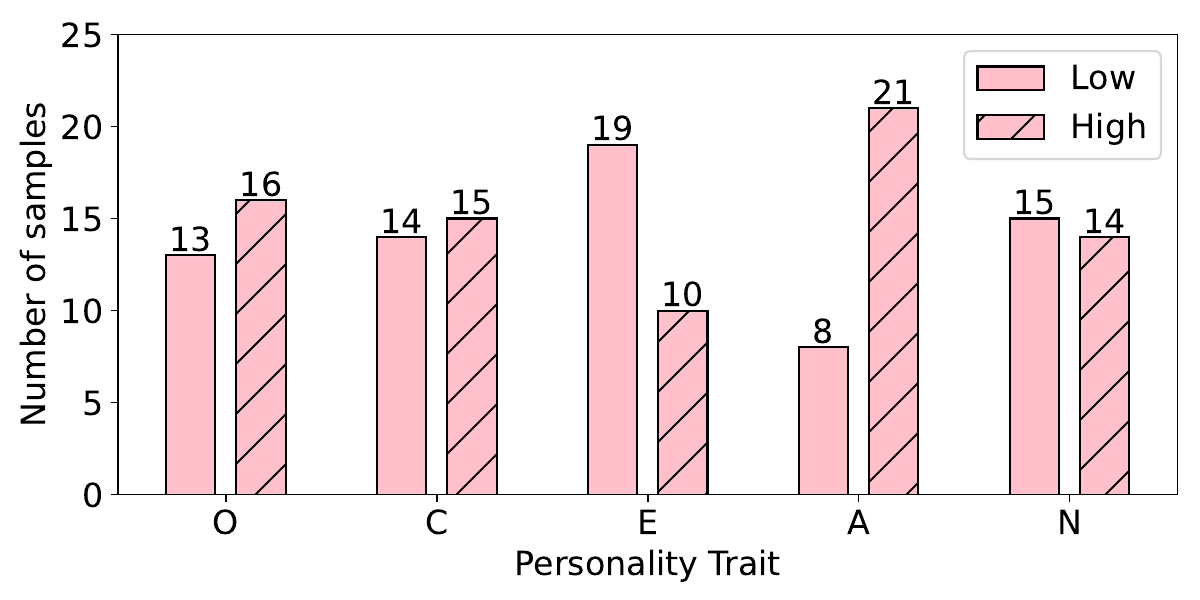}}
	\subfigure[PAN-Testing]{\includegraphics[width=0.32\textwidth]{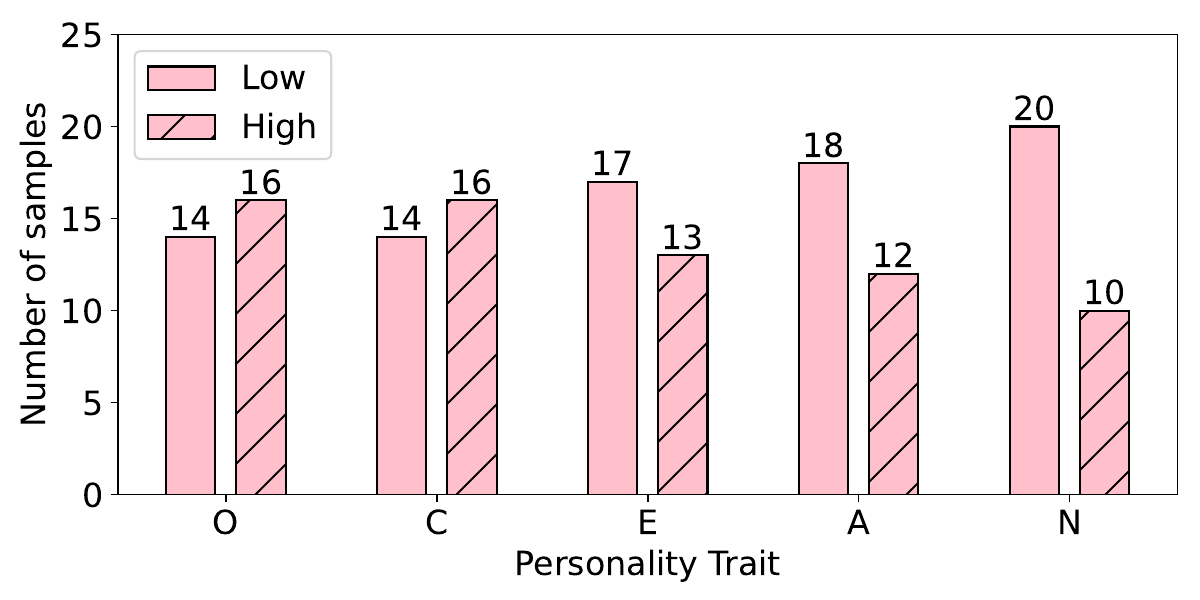}}
	\caption{Statistics of Essays and PAN datasets.}\label{table-dataset}
\end{figure*}

\section{Experiments}\label{sec-experiments}

\subsection{Datasets}\label{sec-experiments-datasets}

We adopt two well-known publicly available datasets in our experiments for text-based Big-Five personality recognition task:

(1) {\bf Essays} \cite{pennebaker1999linguistic}: This stream-of-consciousness dataset consists of 2,467 essays written by psychology students, and the Big-Five personality levels (i.e., low and high levels) of the students were acquired through standardized self-report questionnaire.

(2) {\bf PAN}\footnote{https://pan.webis.de/clef15/pan15-web/author-profiling.html}: This dataset comes from the PAN2015 data science competition, which consists of four language sub-datasets (i.e., Dutch, English, Italian, and Spanish). In this work, we choose the English sub-dataset, which contains 294 users' tweets and their Big-Five personality scores. The Big-Five personality scores of the users were obtained by BFI-10 questionnaire \cite{rammstedt2007measuring}. Note that, similar to \cite{butt2020multimodal}, for each of the five personality traits, we adopt the corresponding mean value to convert personality scores into two personality levels (i.e., low and high levels). To be specific, personality score below the corresponding mean value is converted into the low level, while personality score equal to or above the corresponding mean value is converted into the high level.

Similar to \cite{zhu2022lexical}, we randomly split Essays and PAN datasets into training, validation, and testing sets in the proportion of 8:1:1. The statistics of the two datasets are summarized in Figure~\ref{table-dataset}. 

\subsection{Prompting Strategies}\label{sec-experiments-prompting-strategies}

We employ three classic prompting strategies to explore the personality recognition ability of ChatGPT, including \textit{zero-shot prompting}, \textit{zero-shot CoT prompting}, and \textit{one-shot prompting}. The reason for using one-shot prompting alone is that ChatGPT has a limitation on the length of input. Considering that the texts in both Essays and PAN datasets are normally long (i.e., the average lengths of texts in Essays and PAN datasets are 749 and 1,405 respectively), we only provide one demonstration example in the input (i.e., one-shot prompting) without offering more demonstration examples (e.g., two-shot prompting). In addition, inspired by existing NLP research mining valuable text information at different levels (e.g., word level, sentence level, and document level) \cite{chen2016neural,le2014distributed,ezaldeen2022hybrid,ristoski2020large}, we design level-oriented prompting strategy to guide ChatGPT in analyzing text at a specified level. Concretely, we combine the level-oriented prompting strategy with zero-shot CoT prompting to construct \textit{zero-shot level-oriented CoT prompting}. The reason for constructing zero-shot level-oriented CoT prompting based on zero-shot CoT prompting is that ChatGPT with zero-shot CoT prompting has better overall performance on the two datasets when compared to zero-shot prompting and one-shot prompting (see Section~\ref{sec-overall-performance}). Hence, we would like to see whether the level-oriented prompting strategy could further enhance the effectiveness of zero-shot CoT prompting. Note that, the four prompting strategies require ChatGPT to simultaneously output the person's levels of five personality traits (i.e., O, C, E, A, and N) based on given text.

(1) {\bf Zero-Shot prompting}

\textit{Analyze the person-generated text, determine the person's levels of Openness, Conscientiousness, Extraversion, Agreeableness, and Neuroticism. Only return Low or High.}

\textit{Text: "[Text]"}

\textit{Level:}

(2) {\bf Zero-Shot CoT prompting}

\textit{Analyze the person-generated text, determine the person's levels of Openness, Conscientiousness, Extraversion, Agreeableness, and Neuroticism. Only return Low or High.}

\textit{Text: "[Text]"}

\textit{Level: Let's think step by step:}

(3) {\bf One-Shot prompting}

\textit{Analyze the person-generated text, determine the person's levels of Openness, Conscientiousness, Extraversion, Agreeableness, and Neuroticism. Only return Low or High.}

\textit{Text: "[Example Text]"}

\textit{Level: [Openness Level of Example Text] Openness, [Conscientiousness Level of Example Text] Conscientiousness, [Extraversion Level of Example Text] Extraversion, [Agreeableness Level of Example Text] Agreeableness, [Neuroticism Level of Example Text] Neuroticism}

\textit{Text: "[Text]"}

\textit{Level:}

Note that, to minimize the variance resulting from the sampling of demonstration examples, we randomly select three demonstration examples for conducting experiments and reporting the average performance.

(4) {\bf Zero-Shot Level-Oriented CoT prompting}

We modify zero-shot CoT prompting as follow to construct zero-shot level-oriented CoT prompting, while [Specified Level] can be set as word level, sentence level, or document level.

\textit{Analyze the person-generated text from [Specified Level], determine the person's levels of Openness, Conscientiousness, Extraversion, Agreeableness, and Neuroticism. Only return Low or High.}

\textit{Text: "[Text]"}

\textit{Level: Let's think step by step:}

\subsection{Baselines}

Based on our literature research, we choose the following representative models as baselines:

(1) {\bf RNN} \cite{yu2017deep}: uses RNN to generate text representation for recognizing Big-Five personality. In addition, the pre-trained GloVe model \cite{pennington2014glove} is used to initialize the word embeddings.

(2) {\bf RoBERTa} \cite{jiang2020automatic}: fine-tunes pre-trained RoBERTa-Base model and utilizes the representation of [CLS] with a linear layer for personality classification.

(3) {\bf HPMN (BERT)} \cite{zhu2022lexical}: is one of the SOTA personality prediction models, which uses the personality lexicons to incorporate relevant external knowledge for enhancing the semantic meaning of the person-generated text. Its performance on Essays and PAN datasets is quoted from the original paper.

\subsection{Evaluation Metrics}
It can be observed from Figure~\ref{table-dataset} that Essays and PAN datasets maintain class balance across most of the five personality traits. Therefore, we use \textit{Accuracy} (the higher the better) \cite{hossin2015review} as the evaluation metric, which is used to measure the personality classification performance. Besides, to make a more intuitive comparison, we adopt Accuracy Improvement Percentage (AIP) to measure the accuracy improvement percentage of ChatGPT against the SOTA model (i.e., HPMN (BERT)), which is calculated as:
\begin{equation}
AIP=\frac{Accuracy_{testmodel}-Accuracy_{SOTA}}{Accuracy_{SOTA}}*100\%
\end{equation}
where $Accuracy_{SOTA}$ and $Accuracy_{testmodel}$ denote the accuracy of the SOTA model and the test model such as ChatGPT with zero-shot prompting.

\subsection{Implementation Details}
For the usage of ChatGPT, we adopt the representative version of ChatGPT (i.e., gpt-3.5-turbo). In addition, we set the temperature to 0 for producing more deterministic and focused responses. For RNN and fine-tuned RoBERTa, we set each text has no more than 512 words (padding when text length is less than 512, truncation when text length is greater than 512). Besides, for RNN, the dimension of hidden state, the batch size, and the learning rate are set to 128, 32, and 1e-3 respectively. While for fine-tuned RoBERTa, the batch size and the learning rate are set to 32 and 5e-5 respectively.

\begin{table*}
\centering
\caption{The experimental results in terms of classification accuracy on Essays dataset. The boldface indicates the best model results of Essays dataset, and the underline indicates the second best model result of Essays dataset. SOTA stands for HPMN (BERT)}\label{table-overall-result-essays}
\begin{tabular}{llcccccc}
\toprule
\multicolumn{2}{c}{Model}                                                                                                                          & O                                & C                                & E                                & A                                & N                                & \multicolumn{1}{l}{Average}     \\ \midrule
\multicolumn{1}{l}{\multirow{3}{*}{Baseline}}                                                                                 & RNN                & 57.3\%                           & 52.8\%                           & 45.2\%                           & 45.2\%                           & 50.8\%                           & 50.3\%                           \\ 
\multicolumn{1}{l}{}                                                                                                          & RoBERTa            & 64.9\%                           & 52.8\%                           & 51.2\%                           & 58.1\%                           & 59.7\%                           & 57.3\%                           \\ 
\multicolumn{1}{l}{}                                                                                                          & SOTA               & \textbf{81.8\%} & \textbf{79.6\%} & \textbf{81.1\%} & \textbf{80.7\%} & \textbf{81.7\%} & \textbf{80.9\%} \\ \midrule
\multicolumn{1}{l}{\multirow{3}{*}{Classic prompting strategy}}                                                               & ChatGPT$_{ZS}$     & 60.9\%                           & 56.0\%                           & 50.8\%                           & 58.9\%                           & 60.5\%                           & 57.4\%                           \\ 
\multicolumn{1}{l}{}                                                                                                          & ChatGPT$_{CoT}$    & {\ul 65.7\%}    & 53.2\%                           & 49.2\%                           & {\ul 60.9\%}    & 60.1\%                           & 57.8\%                           \\
\multicolumn{1}{l}{}                                                                                                          & ChatGPT$_{OS}$     & 58.4\%                           & 54.5\%                           & {\ul 59.0\%}    & 58.8\%                           & 60.5\%                           & 58.2\%                           \\ \midrule
\multicolumn{1}{l}{\multirow{3}{*}{\begin{tabular}[c]{@{}l@{}}Level-oriented prompting strategy\\ (Our method)\end{tabular}}} & ChatGPT$_{CoT\_W}$ & 59.3\%                           & {\ul 56.5\%}    & 50.4\%                           & 58.9\%                           & {\ul 61.3\%}    & 57.3\%                           \\ 
\multicolumn{1}{l}{}                                                                                                          & ChatGPT$_{CoT\_S}$ & 62.1\%                           & 55.2\%                           & 51.6\%                           & 59.3\%                           & 58.9\%                           & 57.4\%                           \\ 
\multicolumn{1}{l}{}                                                                                                          & ChatGPT$_{CoT\_D}$ & 64.1\%                           & {\ul 56.5\%}    & 51.2\%                           & 59.7\%                           & 60.1\%                           & {\ul 58.3\%}    \\ \midrule
\multicolumn{1}{l}{\multirow{6}{*}{AIP of SOTA}}                                                                              & ChatGPT$_{ZS}$     & -25.6\%                          & -29.6\%                          & -37.4\%                          & -27.0\%                          & -25.9\%                          & -29.0\%                          \\  
\multicolumn{1}{l}{}                                                                                                          & ChatGPT$_{CoT}$    & -19.7\%                          & -33.2\%                          & -39.3\%                          & -24.5\%                          & -26.4\%                          & -28.6\%                          \\ 
\multicolumn{1}{l}{}                                                                                                          & ChatGPT$_{OS}$     & -28.6\%                          & -31.5\%                          & -27.3\%                          & -27.1\%                          & -25.0\%                          & -29.2\%                          \\  
\multicolumn{1}{l}{}                                                                                                          & ChatGPT$_{CoT\_W}$ & -27.5\%                          & -29.0\%                          & -37.9\%                          & -27.0\%                          & -25.0\%                          & -29.2\%                          \\  
\multicolumn{1}{l}{}                                                                                                          & ChatGPT$_{CoT\_S}$ & -24.1\%                          & -30.7\%                          & -36.4\%                          & -26.5\%                          & -27.9\%                          & -29.0\%                          \\ 
\multicolumn{1}{l}{}                                                                                                          & ChatGPT$_{CoT\_D}$ & -21.6\%                          & -29.0\%                          & -36.9\%                          & -26.0\%                          & -26.4\%                          & -27.9\%                          \\ \bottomrule
\end{tabular}
\end{table*}

\begin{table*}
\centering
\caption{The experimental results in terms of classification accuracy on PAN dataset. The boldface indicates the best model results of PAN dataset, and the underline indicates the second best model result of PAN dataset. SOTA stands for HPMN (BERT)}\label{table-overall-result-pan}
\begin{tabular}{llcccccc}
\toprule
\multicolumn{2}{c}{Model}                                                                                                                          & O                                & C                                & E                                & A                                & N                                & \multicolumn{1}{l}{Average}     \\ \midrule
\multicolumn{1}{l}{\multirow{3}{*}{Baseline}}                                                                                 & RNN                & 43.3\%                           & {\ul 60.0\%}    & 33.3\%                           & 43.3\%                           & 56.7\%                           & 47.3\%                           \\  
\multicolumn{1}{l}{}                                                                                                          & RoBERTa            & {\ul 63.3\%}    & 53.3\%                           & 53.3\%                           & 40.0\%                           & {\ul 66.7\%}    & 55.3\%                           \\  
\multicolumn{1}{l}{}                                                                                                          & SOTA               & \textbf{66.8\%} & \textbf{64.6\%} & 68.8\%                           & 66.3\%                           & \textbf{71.3\%} & \textbf{67.5\%} \\ \midrule
\multicolumn{1}{l}{\multirow{3}{*}{Classic prompting strategy}}                                                               & ChatGPT$_{ZS}$     & 50.0\%                           & 50.0\%                           & 66.7\%                           & \textbf{70.0\%} & 50.0\%                           & 57.3\%                           \\  
\multicolumn{1}{l}{}                                                                                                          & ChatGPT$_{CoT}$    & 60.0\%                           & 50.0\%                           & {\ul 70.0\%}    & {\ul 66.7\%}    & 56.7\%                           & 60.7\%                           \\ 
\multicolumn{1}{l}{}                                                                                                          & ChatGPT$_{OS}$     & 46.7\%                           & 42.2\%                           & 54.4\%                           & 57.8\%                           & 45.6\%                           & 49.3\%                           \\ \midrule
\multicolumn{1}{l}{\multirow{3}{*}{\begin{tabular}[c]{@{}l@{}}Level-oriented prompting strategy\\ (Our method)\end{tabular}}} & ChatGPT$_{CoT\_W}$ & {\ul 63.3\%}    & 53.3\%                           & 66.7\%                           & 63.3\%                           & 56.7\%                           & 60.7\%                           \\
\multicolumn{1}{l}{}                                                                                                          & ChatGPT$_{CoT\_S}$ & 60.0\%                           & 56.7\%                           & \textbf{73.3\%} & \textbf{70.0\%} & 53.3\%                           & {\ul 62.7\%}    \\ 
\multicolumn{1}{l}{}                                                                                                          & ChatGPT$_{CoT\_D}$ & {\ul 63.3\%}    & 46.7\%                           & {\ul 70.0\%}    & {\ul 66.7\%}    & 50.0\%                           & 59.3\%                           \\ \midrule
\multicolumn{1}{l}{\multirow{6}{*}{AIP of SOTA}}                                                                              & ChatGPT$_{ZS}$     & -25.1\%                          & -22.6\%                          & -3.1\%                           & -5.6\%                           & -29.9\%                          & -15.1\%                          \\  
\multicolumn{1}{l}{}                                                                                                          & ChatGPT$_{CoT}$    & -10.2\%                          & -22.6\%                          & +1.7\%                           & +0.6\%                           & -20.5\%                          & -10.1\%                          \\ 
\multicolumn{1}{l}{}                                                                                                          & ChatGPT$_{OS}$     & -30.1\%                          & -34.7\%                          & -20.9\%                          & -12.8\%                          & -36.0\%                          & -27.0\%                          \\  
\multicolumn{1}{l}{}                                                                                                          & ChatGPT$_{CoT\_W}$ & -5.2\%                           & -17.5\%                          & -3.1\%                           & -4.5\%                           & -20.5\%                          & -10.1\%                          \\ 
\multicolumn{1}{l}{}                                                                                                          & ChatGPT$_{CoT\_S}$ & -10.2\%                          & -12.2\%                          & +6.5\%                           & +5.6\%                           & -25.2\%                          & -7.1\%                           \\ 
\multicolumn{1}{l}{}                                                                                                          & ChatGPT$_{CoT\_D}$ & -5.2\%                           & -27.7\%                          & +1.7\%                           & +0.6\%                           & -29.9\%                          & -12.1\%                          \\ \bottomrule
\end{tabular}
\end{table*}

\subsection{Overall Performance ({\bf RQ1})}\label{sec-overall-performance}

Considering that ChatGPT may refuse personality recognition due to some reasons\footnote{One unexpected response of ChatGPT: ``Unfortunately, there is not enough information in the provided text to accurately determine the person's levels of Openness, Conscientiousness, Extraversion, Agreeableness, and Neuroticism.".}, we adopt \textit{Majority} approach to obtain the prediction results when encountering such rare situations. Specifically, for each personality trait, we regard the majority personality level in training set as the personality level of each sample in testing set. The experimental results on Essays and PAN datasets are shown in Table~\ref{table-overall-result-essays} and Table~\ref{table-overall-result-pan}. Concretely, ChatGPT$_{ZS}$, ChatGPT$_{CoT}$, and ChatGPT$_{OS}$ represent ChatGPT with zero-shot prompting, zero-shot CoT prompting, and one-shot prompting. In addition, ChatGPT$_{CoT\_W}$, ChatGPT$_{CoT\_S}$, and ChatGPT$_{CoT\_D}$ denotes ChatGPT with zero-shot level-oriented CoT prompting, while [Specified Level] is set to word level, sentence level, and document level respectively.

{\bf Results of zero-shot prompting}. As shown in Table~\ref{table-overall-result-essays} and Table~\ref{table-overall-result-pan}, ChatGPT$_{ZS}$ has better performance than the traditional neural network RNN on both Essays and PAN datasets. For example, relative to RNN, ChatGPT$_{ZS}$ increases its average classification accuracy from 50.3\% to 57.4\% on Essays dataset. Furthermore, ChatGPT$_{ZS}$ not only performs comparably to fine-tuned RoBERTa on Essays dataset (e.g., 57.4\% vs. 57.3\% in terms of average classification accuracy) but also outperforms fine-tuned RoBERTa on PAN dataset (e.g., 57.3\% vs. 55.3\% w.r.t. average classification accuracy). Therefore, ChatGPT$_{ZS}$ exhibits incredible text-based personality recognition ability under zero-shot setting. Since the SOTA model is a task-specific fully-supervised model with complex architecture for personality recognition task, the performance of ChatGPT$_{ZS}$ falls far behind that of the SOTA model on the two datasets (e.g., 57.3\% vs. 67.5\% w.r.t. average classification accuracy on PAN dataset). However, another interesting observation is that compared with Essays dataset (i.e., the relatively large-scale dataset), ChatGPT$_{ZS}$ shows a relatively higher AIP on PAN dataset (i.e., the relatively small-scale dataset). For example, the AIP of ChatGPT$_{ZS}$ against the SOTA model on Essays and PAN datasets are -29.0\% and -15.1\% respectively. Furthermore, ChatGPT$_{ZS}$ even surpasses the SOTA model when predicting personality trait $A$ on PAN dataset (i.e., 70.0\% vs. 66.3\%). The possible reason is that PAN dataset provides relatively fewer training data for the fully-supervised SOTA model, preventing it from fully learning the differences in personality levels. In contrast, ChatGPT$_{ZS}$ does not require training data and relies solely on its existing knowledge under zero-shot setting, narrowing the performance gap between ChatGPT$_{ZS}$ and the SOTA model.

\begin{figure*}
	\centering
	\subfigure[One output of ChatGPT$_{ZS}$]{\label{figure-chatgpt-zs-output}\includegraphics[width=0.8\textwidth]{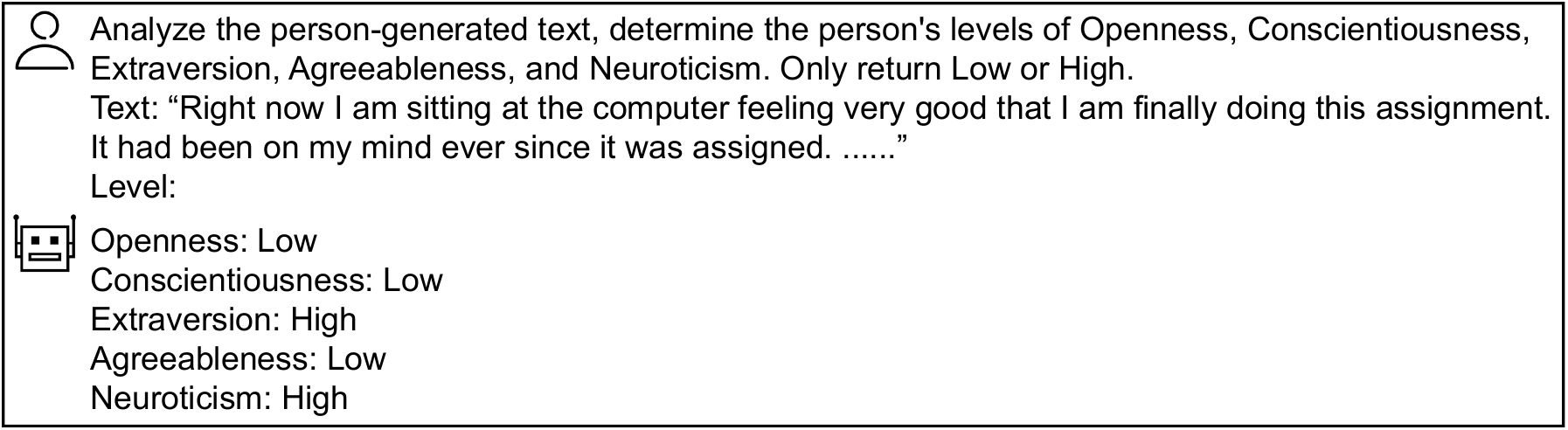}}

	\subfigure[One output of ChatGPT$_{CoT}$]{\label{figure-chatgpt-cot-output}\includegraphics[width=0.8\textwidth]{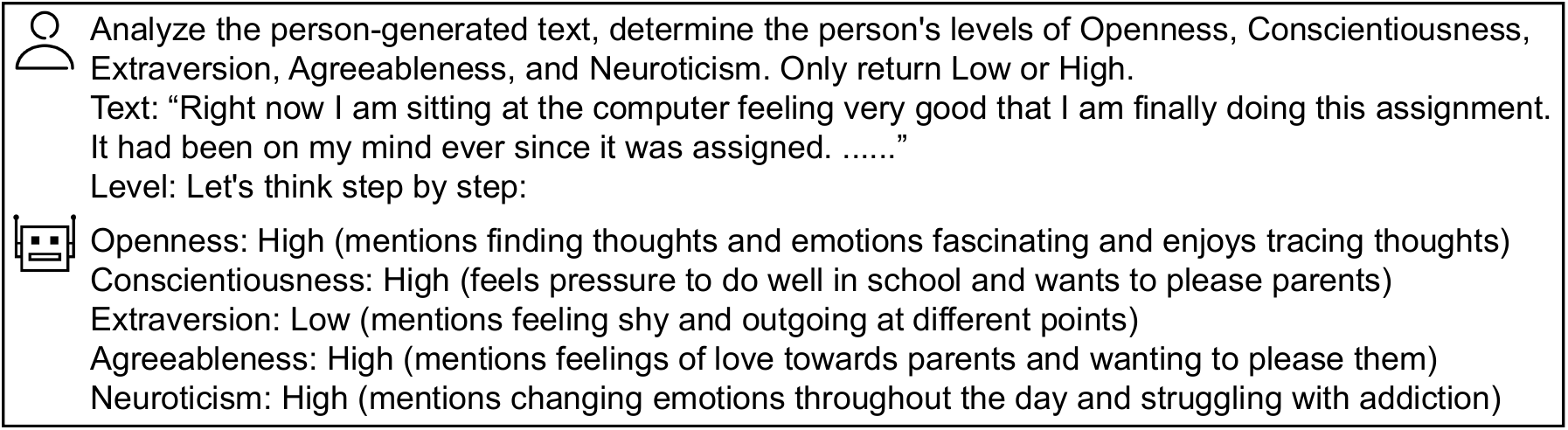}}
	\caption{Examples of ChatGPT$_{ZS}$'s output and ChatGPT$_{CoT}$'s output.}\label{figure-chatgpt-output}
\end{figure*} 

\begin{figure}
	\centering
	\subfigure[None]{\label{figure-chatgpt-cot-explain-none}\includegraphics[width=0.2\textwidth]{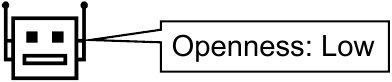}}

	\subfigure[Original Content]{\label{figure-chatgpt-cot-explain-original-content}\includegraphics[width=0.45\textwidth]{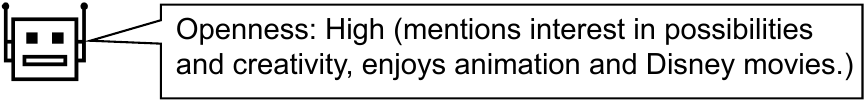}}

	\subfigure[Logical Reasoning]{\label{figure-chatgpt-cot-explain-logical-reasoning}\includegraphics[width=0.45\textwidth]{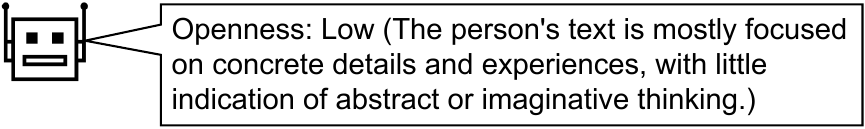}}
	\caption{Examples of three types of natural language explanations generated by ChatGPT$_{CoT}$ for recognizing personality trait $O$.}\label{figure-chagpt-cot-explain}
\end{figure}

\begin{figure}
	\centering
	\subfigure[Essays]{\label{figure-chatgpt-cot-explain-essays}\includegraphics[width=0.38\textwidth]{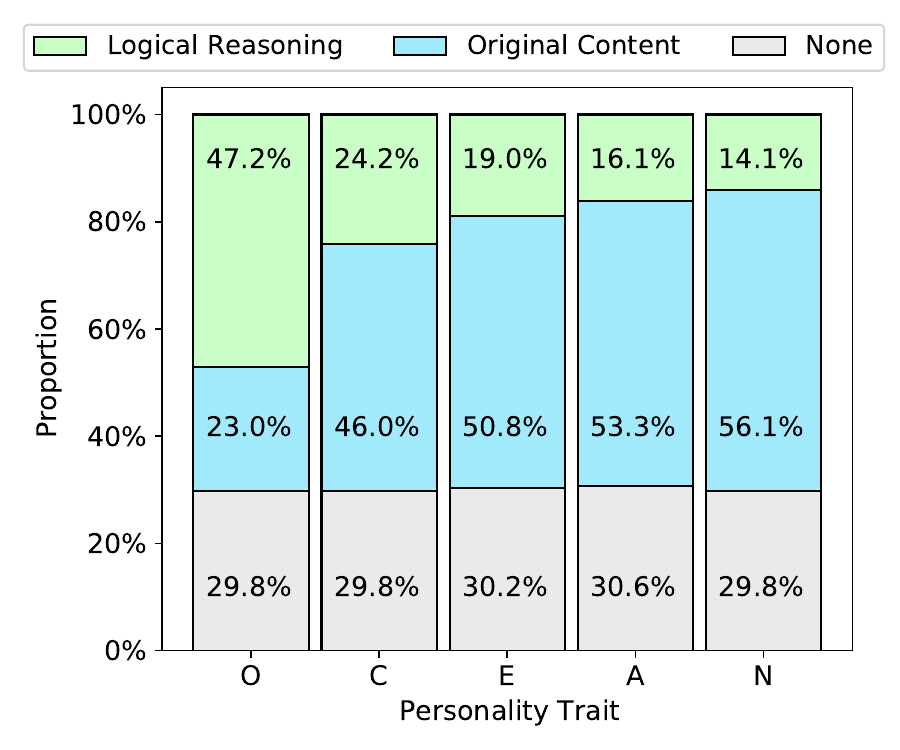}}

	\subfigure[PAN]{\label{figure-chatgpt-cot-explain-pan}\includegraphics[width=0.38\textwidth]{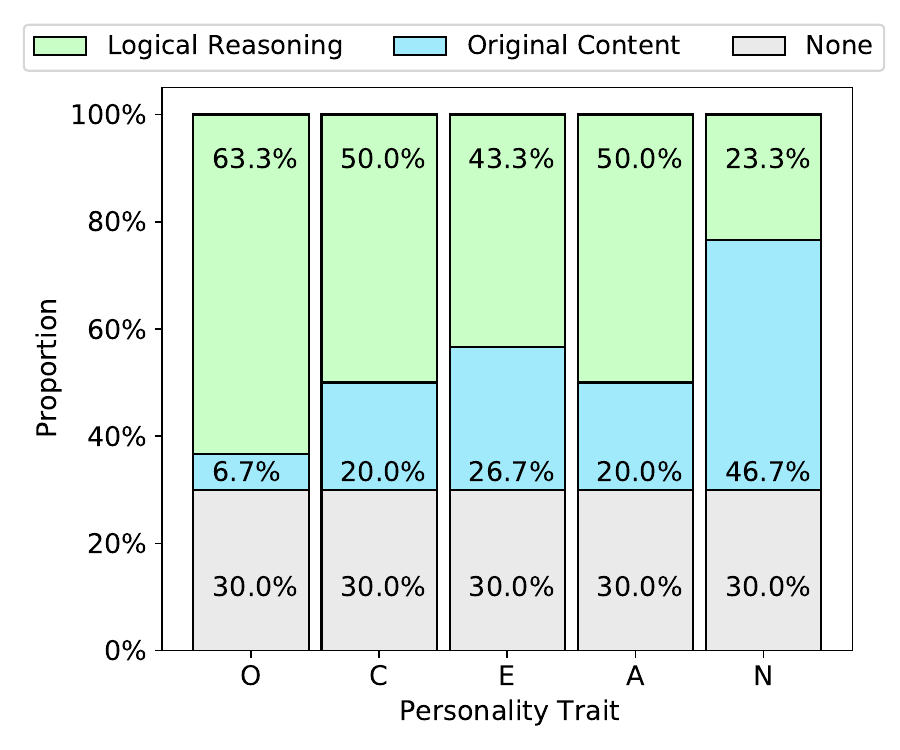}}
	\caption{Distribution of three types of explanations on different datasets and personality traits.}\label{figure-chagpt-cot-explain-distribution}
\end{figure}

{\bf Results of zero-shot CoT prompting}. Table~\ref{table-overall-result-essays} and Table~\ref{table-overall-result-pan} reveal that zero-shot CoT prompting could effectively enhance ChatGPT's ability on text-based personality recognition task. For example, ChatGPT$_{CoT}$ increases its average classification accuracy from 57.3\% to 60.7\% on PAN dataset when compared with ChatGPT$_{ZS}$. As for reason, with the help of zero-shot CoT prompting, ChatGPT$_{CoT}$ can perform more complex logical reasoning, so as to accurately complete the personality prediction task. Besides, ChatGPT$_{ZS}$ only provides final prediction results (see Figure~\ref{figure-chatgpt-zs-output}), while ChatGPT$_{CoT}$ could provide additional natural language explanations for its prediction results in most cases (see Figure~\ref{figure-chatgpt-cot-output}). The natural language explanations generated by ChatGPT$_{CoT}$ not only enhance users' trust in the prediction results but also enables developers to obtain a better understanding of the knowledge deficiencies in ChatGPT. To gain a deep insight into the natural language explanations generated by ChatGPT$_{CoT}$, we categorize the nature language explanations into three types: (1) \textit{None}: no explanation or refuse personality recognition; (2) \textit{Original Content}: only the original text is provided as explanation; (3) \textit{Logical Reasoning}: logical reasoning based on the original text. Figure~\ref{figure-chagpt-cot-explain} shows the examples of three types of natural language explanations for the prediction of personality trait $O$, and Figure~\ref{figure-chagpt-cot-explain-distribution} illustrates the distribution of three types of natural language explanations on different datasets and personality traits. As depicted in Figure~\ref{figure-chagpt-cot-explain-distribution}, on both Essays and PAN datasets, ChatGPT$_{CoT}$ provides more natural language explanations of the logical reasoning type for the prediction of personality trait $O$, while offering more natural language explanations of the original content type when identifying personality trait $N$. With regard to possible reasons, personality trait $O$ reflects whether a person is creative/open-minded (with high level) or reflective/conventional (with low level) \cite{digman1990personality}, which may not be directly presented in person-generated text. Hence, the prediction of personality trait $O$ requires ChatGPT to engage in more logical reasoning for a deeper analysis of given text. For example, as shown in Figure~\ref{figure-chatgpt-cot-explain-logical-reasoning}, based on given text, ChatGPT$_{CoT}$ infers that \textit{the person's text is mostly focused on concrete details and experiences, with little indication of abstract or imaginative thinking}. Therefore, ChatGPT$_{CoT}$ predicts that the person has low $O$. On the contrary, personality trait $N$ reflects whether a person is emotionally stable (with low level) or emotionally unstable (with high level) \cite{digman1990personality}. Since individuals normally directly express their negative emotions (e.g., anxiety) in their texts, it is relatively easier for ChatGPT$_{CoT}$ to predict personality trait $N$ based on the original text without logical reasoning. For example, one of natural language explanation of the original content type generated by ChatGPT$_{CoT}$ for predicting personality trait $N$ is \textit{mentions feeling stressed, tense, and worried about health problems and homework overload}. Furthermore, as demonstrated in Figure~\ref{figure-chagpt-cot-explain-distribution}, compared with Essays dataset, ChatGPT$_{CoT}$ provides relatively more natural language explanations of the logical reasoning type for personality recognition on PAN dataset. The possible reason is that Essays dataset consists of stream-of-consciousness essays written by psychology students under professional guidance, while PAN dataset is composed of tweets written freely by various internet users. Hence, compared with the texts in Essays dataset, the texts in PAN datasets generally contain relatively less valuable information, which increases the difficulty of text-based personality prediction on PAN dataset. Therefore, compared to Essays dataset, ChatGPT$_{CoT}$ needs to perform more logical reasoning to accomplish personality recognition task accurately on PAN dataset.

{\bf Results of one-shot prompting}. From Table~\ref{table-overall-result-essays} and Table~\ref{table-overall-result-pan}, it can be observed that by providing a demonstration example, ChatGPT's performance has improved on Essays dataset but largely declined on PAN dataset. To be specific, ChatGPT$_{OS}$ increases its average classification accuracy from 57.4\% to 58.2\% on Essays dataset when compared with ChatGPT$_{ZS}$. However, relative to ChatGPT$_{ZS}$, ChatGPT$_{OS}$ decreases its average classification accuracy from 57.3\% to 49.3\% on PAN dataset. Regarding possible reasons, on the one hand, as mentioned above, the texts in Essays dataset generally contain more valuable information when compared to PAN dataset. Hence, there is a higher probability of selecting samples containing more invalid information from PAN dataset than from Essays dataset, thereby affecting ChatGPT$_{OS}$'s learning of the relationship between text and Big-Five personality on PAN dataset. On the other hand, the persons in Essays dataset are all psychology students, while the persons in PAN dataset are various internet users from different age groups (from 18 years old to over 50 years old). Hence, without the corresponding demographic attributes (e.g., age) provided, the demonstration example selected from the training set of PAN dataset may not assist ChatGPT$_{OS}$ in predicting the personalities of certain groups. For instance, if the demonstration example is generated by a young person, the association between text and personality that ChatGPT$_{OS}$ learns from this demonstration example may not be helpful in predicting the personality of an old person.

{\bf Results of zero-shot level-oriented prompting}. Table~\ref{table-overall-result-essays} and Table~\ref{table-overall-result-pan} demonstrate that guiding ChatGPT$_{CoT}$ to analyze given text from specified level could help ChatGPT in analyzing given text more targeted and completing personality prediction task precisely. For example, by guiding ChatGPT$_{CoT\_D}$ to analyze given text from document level, its performance on Essays dataset can rival the performance of ChatGPT$_{OS}$ (58.3\% vs. 58.2\% w.r.t. average classification accuracy). Similarly, on PAN dataset, when ChatGPT$_{CoT\_S}$ is guided to analyze given text from sentence level, its average classification accuracy has been a notable improvement when compared to ChatGPT$_{CoT}$, rising from 57.3\% to 62.7\%. We believe the possible reason is that the texts in Essays dataset were written within a limited time frame, making it more suitable for conducting overall analysis from document level. On the other hand, the texts in PAN dataset are composed of tweets posted at different times. Hence, it is more appropriate to analyze given text in PAN dataset from sentence level, which is helpful to mine diverse individual information reflected in different tweets. This discovery not only helps optimize existing promptings for text analysis but also offers new insights into eliciting various abilities of LLMs in a fine-grained manner.

\begin{table*}
\centering
\caption{The distribution of different demographic attributes in PAN dataset}\label{table-attribute-distribution}
\begin{tabular}{lccl}
\toprule
Demographic Attribute          & \multicolumn{2}{c}{Distribution}         & {[}Corresponding Attribute{]} \\ \midrule
\multirow{2}{*}{Gender} & \multicolumn{1}{l}{Man}      & 174 (50\%) & a man                         \\ 
                        & \multicolumn{1}{c}{Woman}    & 174 (50\%) & a woman                       \\ \midrule
\multirow{4}{*}{Age}    & \multicolumn{1}{c}{18 to 24} & 114 (39\%) & aged between 18 and 24        \\  
                        & \multicolumn{1}{c}{25 to 34} & 118 (40\%) & aged between 25 and 34        \\ 
                        & \multicolumn{1}{c}{35 to 49} & 42 (14\%)  & aged between 35 and 49        \\ 
                        & \multicolumn{1}{c}{$\geq$50}       & 20 (7\%)   & aged 50 or over               \\ \bottomrule
\end{tabular}
\end{table*}

\begin{figure*}
	\centering
	\subfigure[O]{\includegraphics[width=0.32\textwidth]{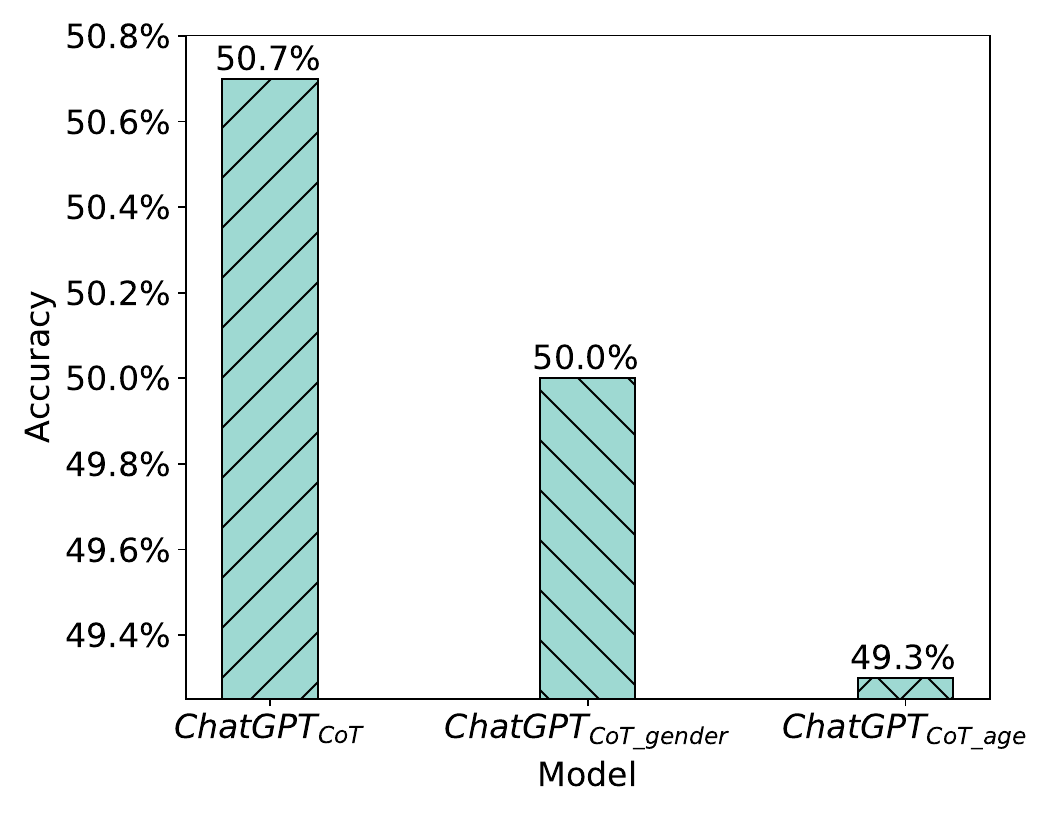}}
	\subfigure[C]{\includegraphics[width=0.32\textwidth]{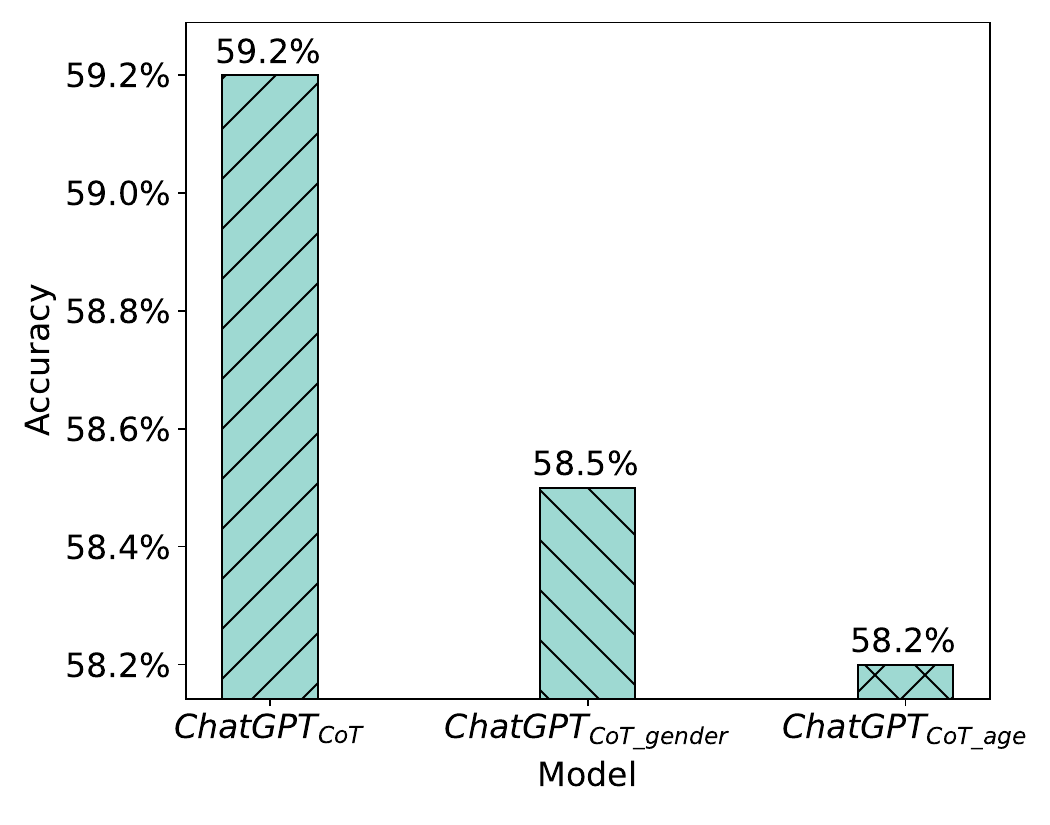}}
	\subfigure[E]{\includegraphics[width=0.32\textwidth]{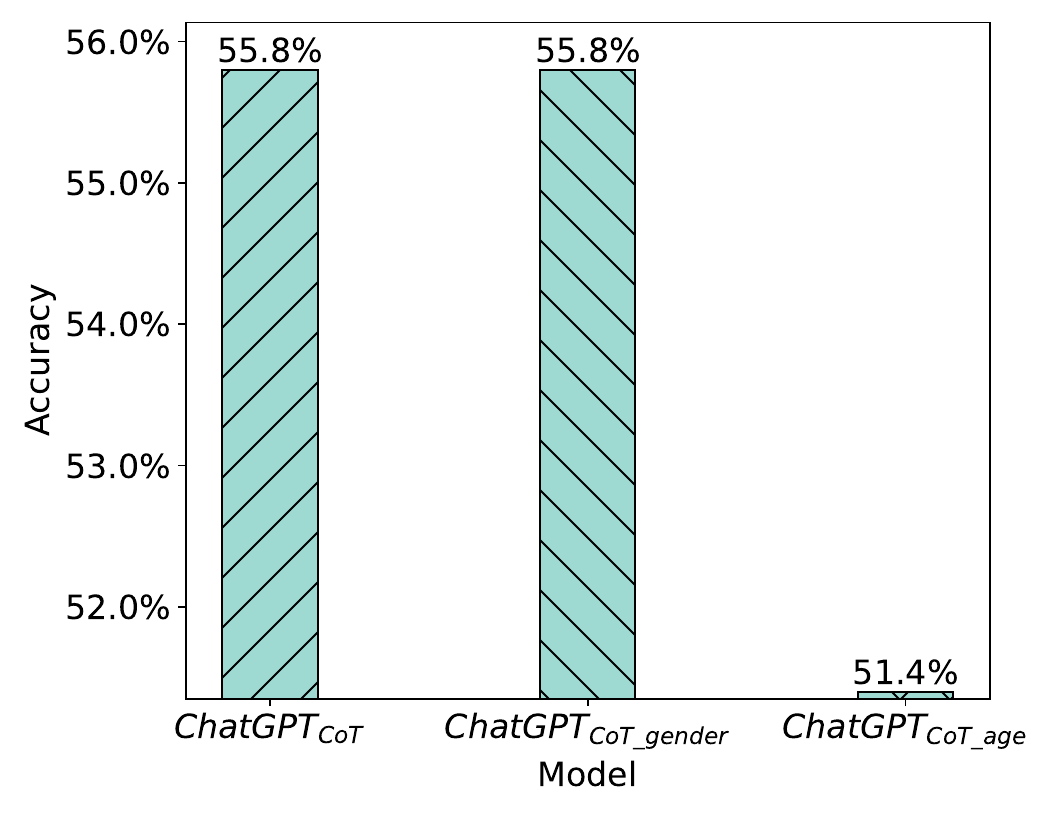}}
	\subfigure[A]{\includegraphics[width=0.32\textwidth]{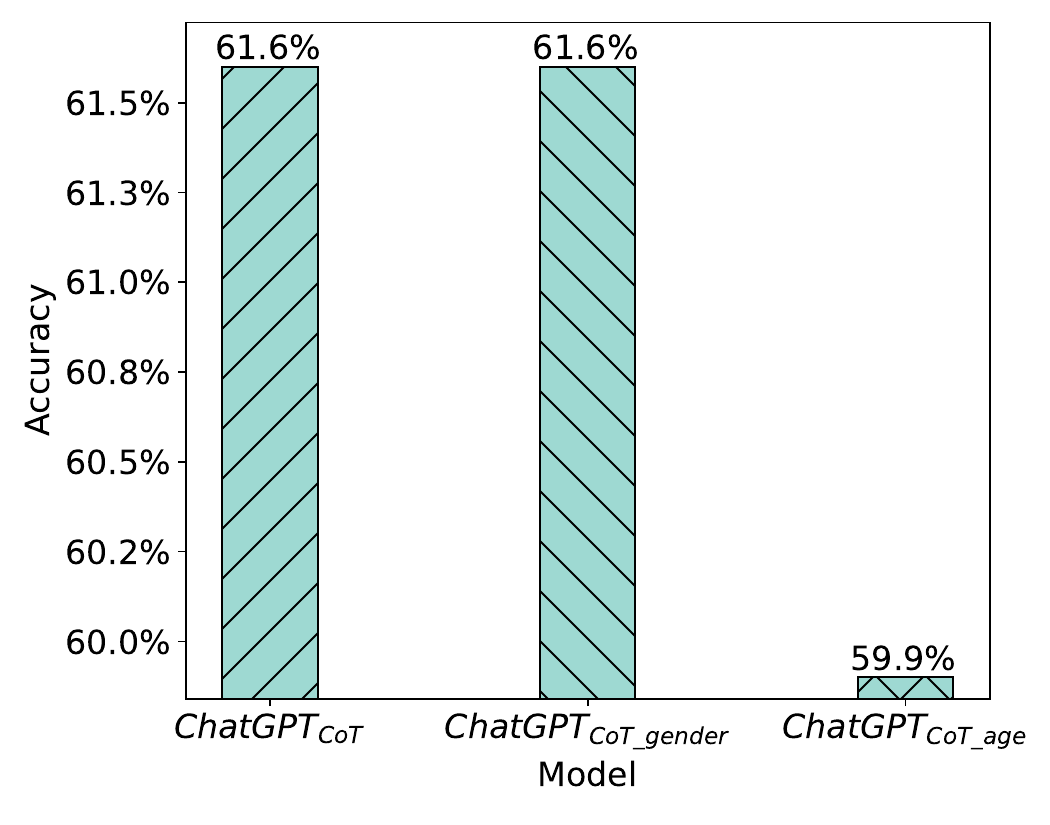}}
	\subfigure[N]{\includegraphics[width=0.32\textwidth]{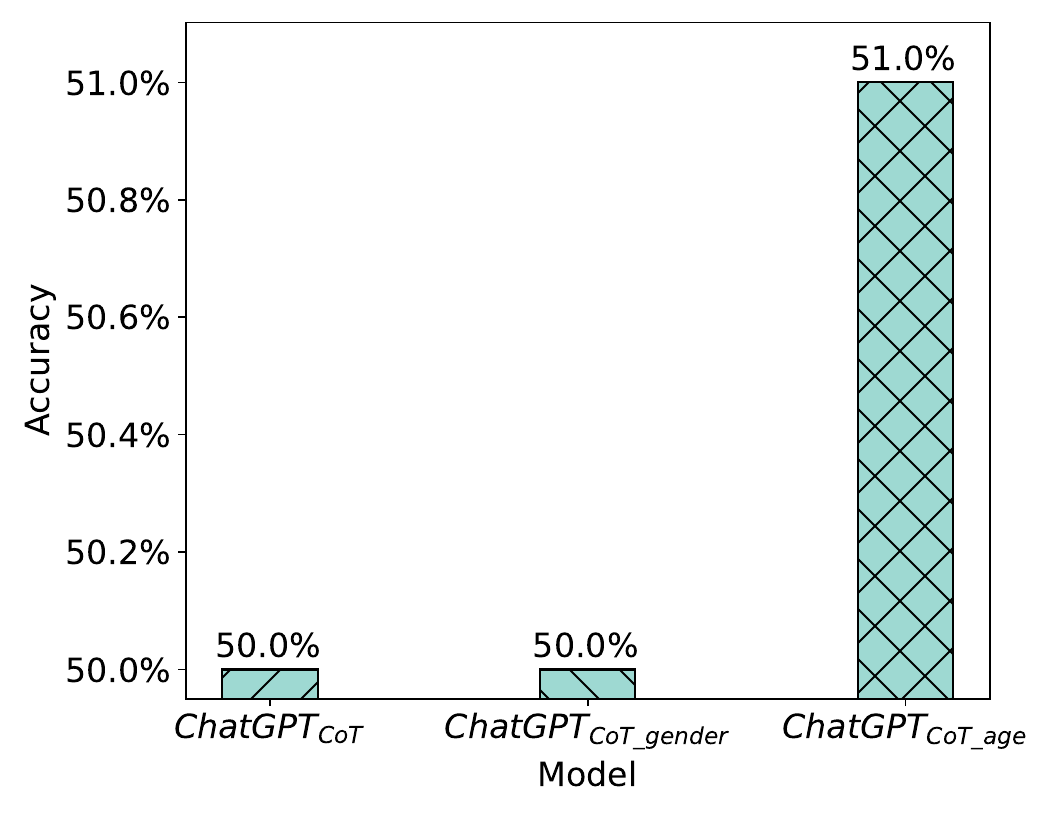}}
	\subfigure[Average]{\includegraphics[width=0.32\textwidth]{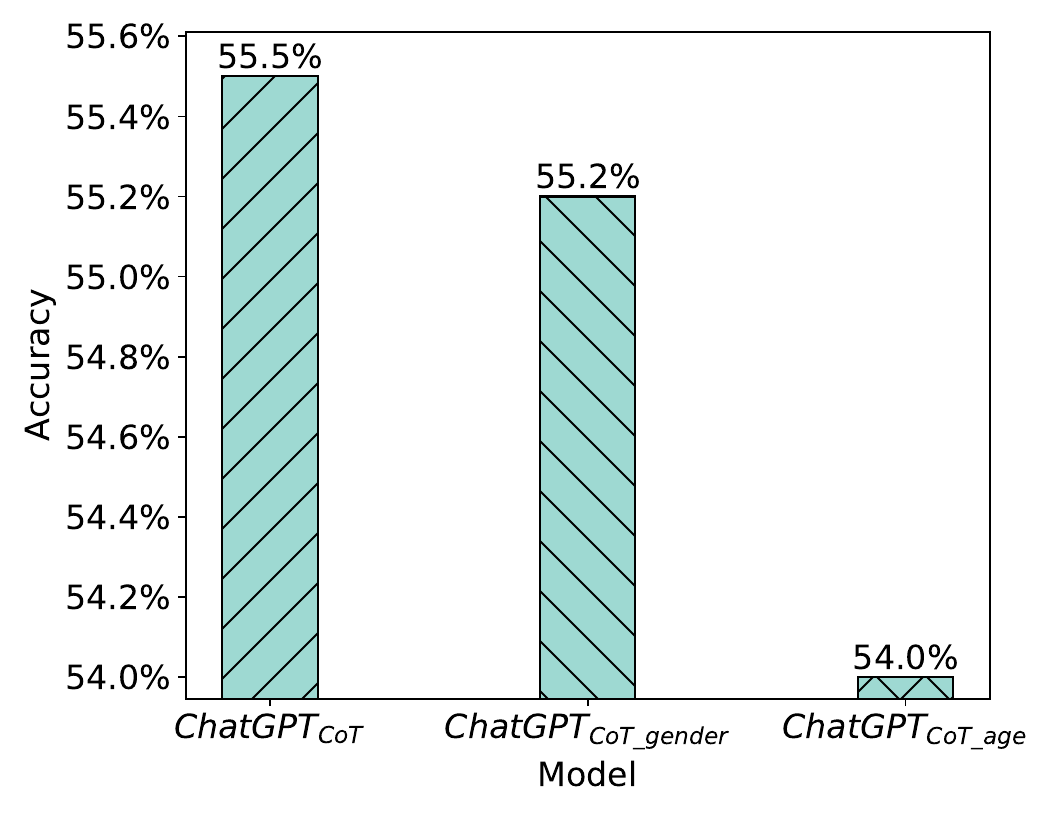}}
	\caption{The experimental results of ChatGPT$_{CoT}$, ChatGPT$_{CoT\_gender}$, and ChatGPT$_{CoT\_age}$ on PAN dataset}\label{figure-attribute-results}
\end{figure*} 

\subsection{Fairness of ChatGPT on Personality Recognition ({\bf RQ2})}\label{section-fairness}

Considering that LLMs may be unfair to certain groups due to social bias in its large pre-training corpus \cite{zhang2023chatgpt}, we further investigate the fairness of ChatGPT on personality prediction task across different groups. To be specific, we adopt ChatGPT$_{CoT}$ with different demographic attributes for personality prediction on PAN dataset, as PAN dataset provides various demographic attributes, including gender and age (see Table~\ref{table-attribute-distribution}). Concretely, we modify zero-shot CoT prompting as follow to provide ChatGPT with specific demographic attribute corresponding to given text:

\textit{Analyze the person-generated text, determine the person's level of Openness, Conscientiousness, Extraversion, Agreeableness, and Neuroticism. Only return Low or High. Note that, the person is [Corresponding Attribute].}

\textit{Text: "[Text]"}

\textit{Level: Let's think step by step:}

Please refer to Table~\ref{table-attribute-distribution} for the setting of [Corresponding Attribute]. For example, [Corresponding Attribute] is set to \textit{aged between 18 and 24} when the age of the corresponding person is between 18 and 24 years old. To be specific, ChatGPT$_{CoT\_gender}$ and ChatGPT$_{CoT\_age}$ represent ChatGPT with the modified zero-shot CoT promptings, which incorporates gender and age information respectively.

\begin{figure*}
	\centering
	\subfigure[Gender (Woman) - O]{\label{figure-distribution-prediction-gender-woman-O}\includegraphics[width=0.32\textwidth]{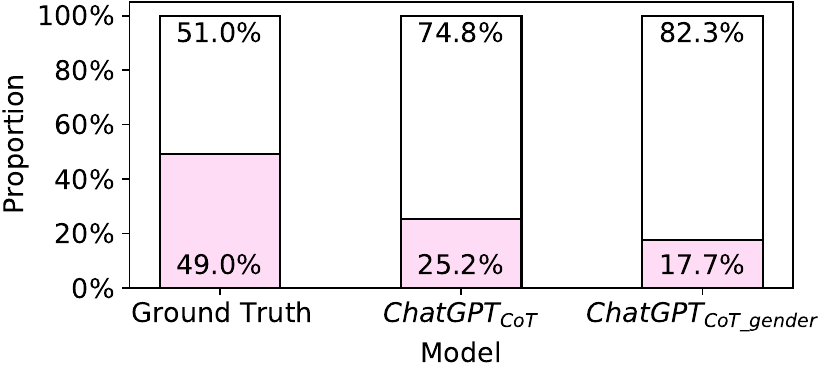}}
	\subfigure[Gender (Woman) - C]{\includegraphics[width=0.32\textwidth]{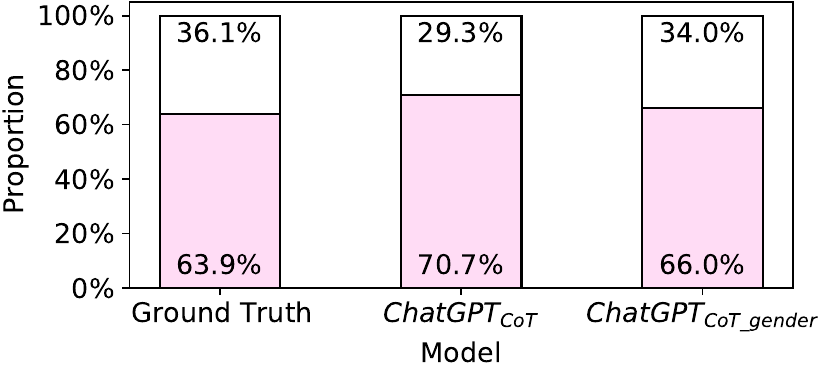}}
	\subfigure[Gender (Woman) - E]{\includegraphics[width=0.32\textwidth]{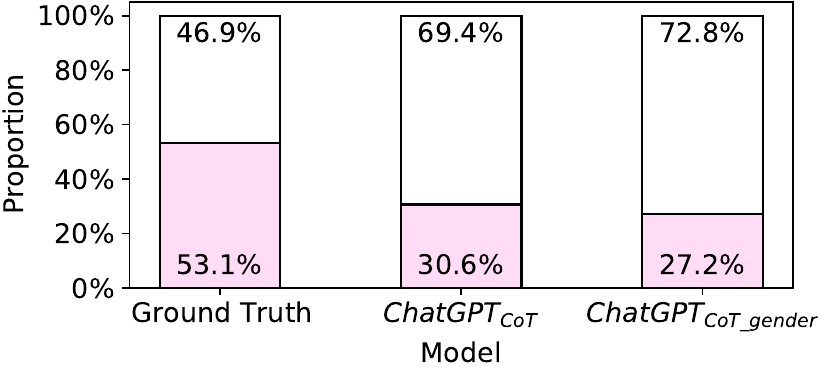}}
	\subfigure[Gender (Woman) - A]{\includegraphics[width=0.32\textwidth]{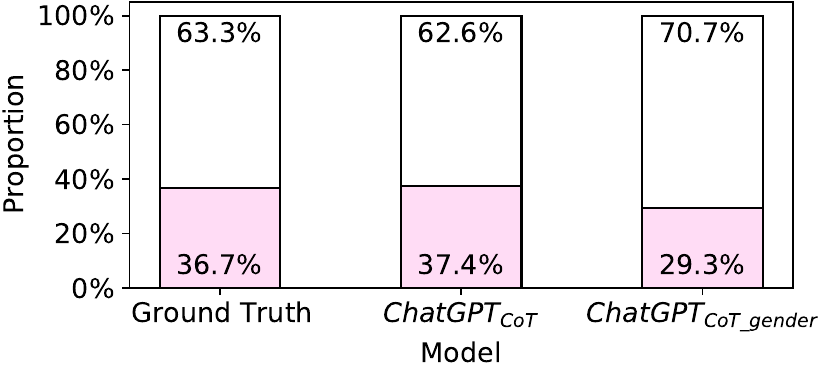}}
	\subfigure[Gender (Woman) - N]{\includegraphics[width=0.32\textwidth]{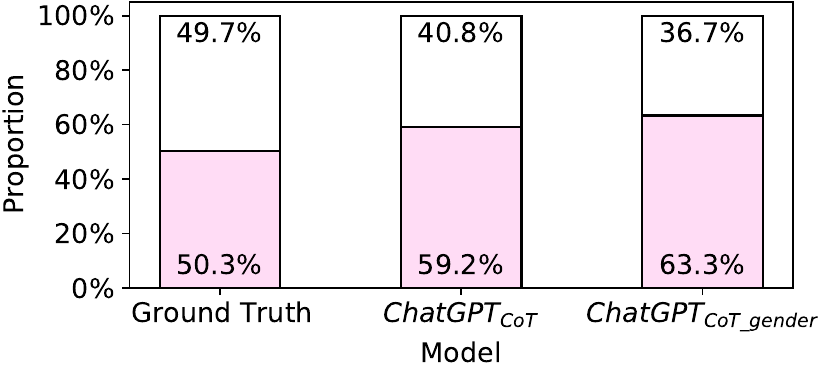}}
	\caption{Distribution of prediction results of ChatGPT$_{CoT}$ and ChatGPT$_{CoT\_gender}$ towards woman group, while purple and white denote low and high levels respectively.}\label{figure-distribution-prediction-gender-woman}
\end{figure*} 

\begin{figure*}
	\centering
	\subfigure[Gender (Man) - O]{\label{figure-distribution-prediction-gender-man-O}\includegraphics[width=0.32\textwidth]{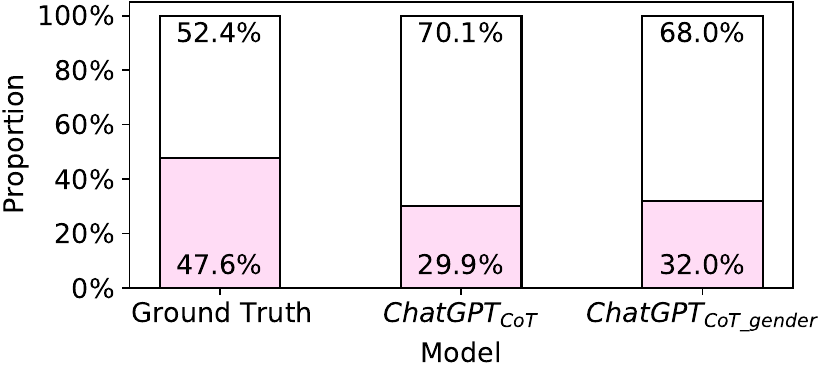}}
	\subfigure[Gender (Man) - C]{\includegraphics[width=0.32\textwidth]{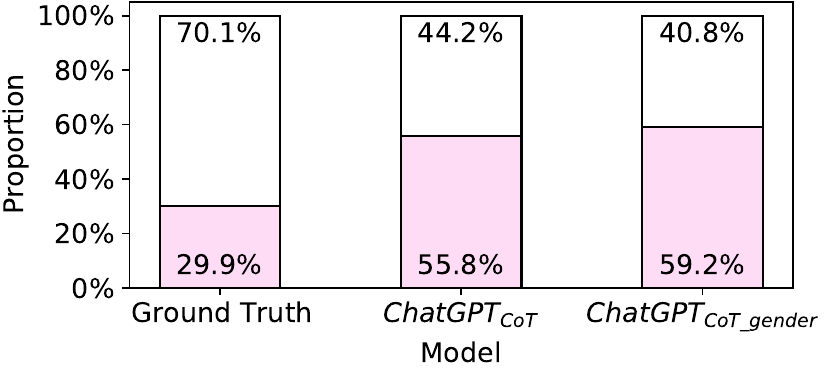}}
	\subfigure[Gender (Man) - E]{\includegraphics[width=0.32\textwidth]{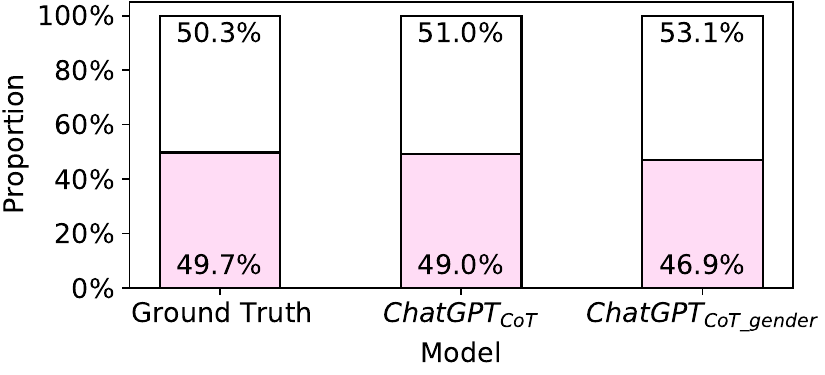}}
	\subfigure[Gender (Man) - A]{\includegraphics[width=0.32\textwidth]{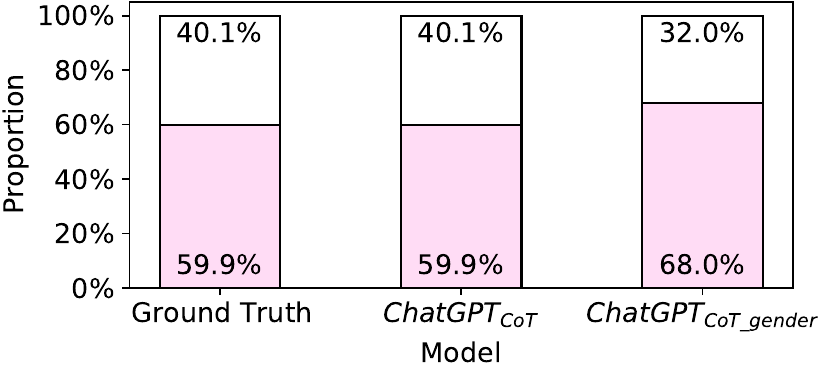}}
	\subfigure[Gender (Man) - N]{\includegraphics[width=0.32\textwidth]{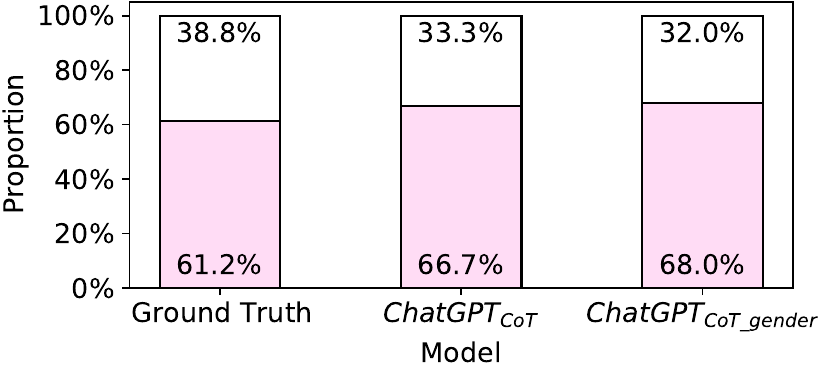}}
	\caption{Distribution of prediction results of ChatGPT$_{CoT}$ and ChatGPT$_{CoT\_gender}$ towards man group, while purple and white denote low and high levels respectively.}\label{figure-distribution-prediction-gender-man}
\end{figure*} 

\begin{figure*}
	\centering
	\subfigure[Age (18 to 24) - O]{\includegraphics[width=0.32\textwidth]{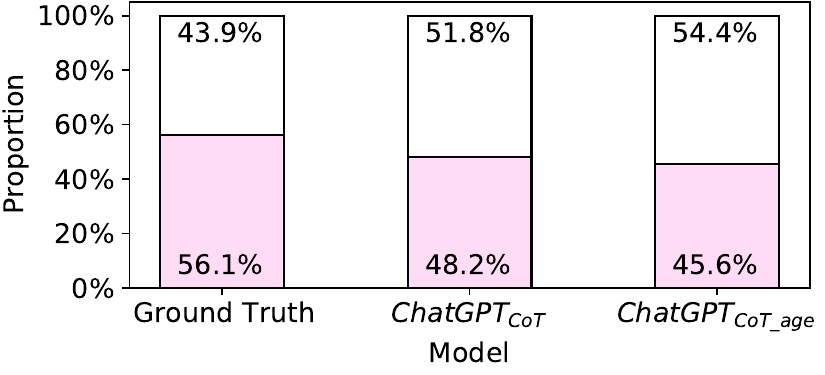}}
	\subfigure[Age (18 to 24) - C]{\includegraphics[width=0.32\textwidth]{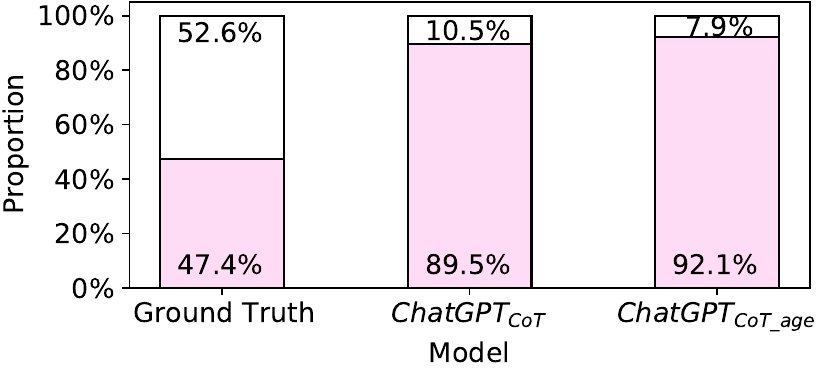}}
	\subfigure[Age (18 to 24) - E]{\includegraphics[width=0.32\textwidth]{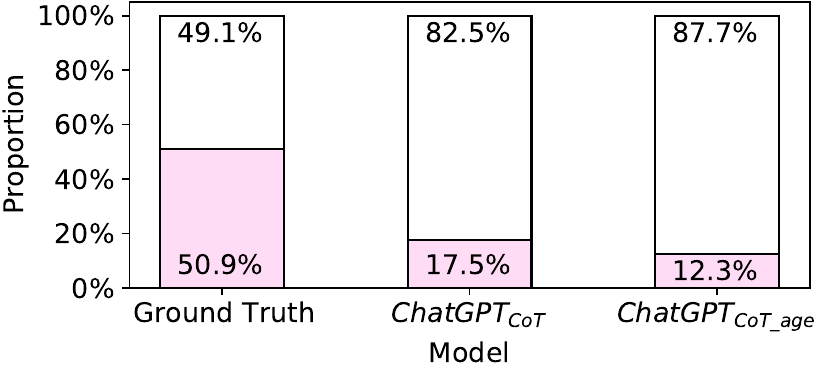}}
	\subfigure[Age (18 to 24) - A]{\includegraphics[width=0.32\textwidth]{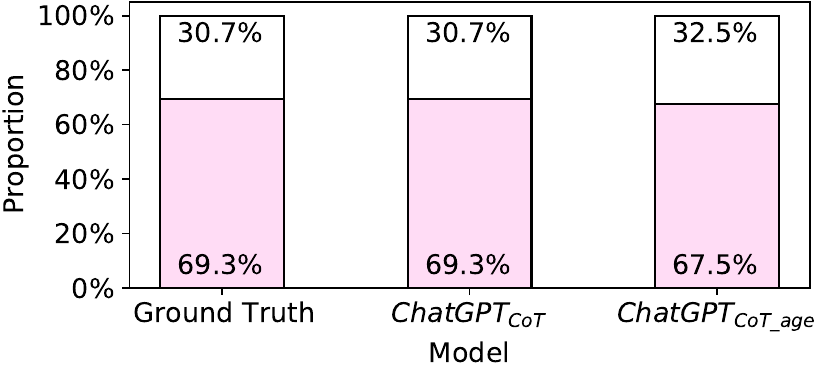}}
	\subfigure[Age (18 to 24) - N]{\includegraphics[width=0.32\textwidth]{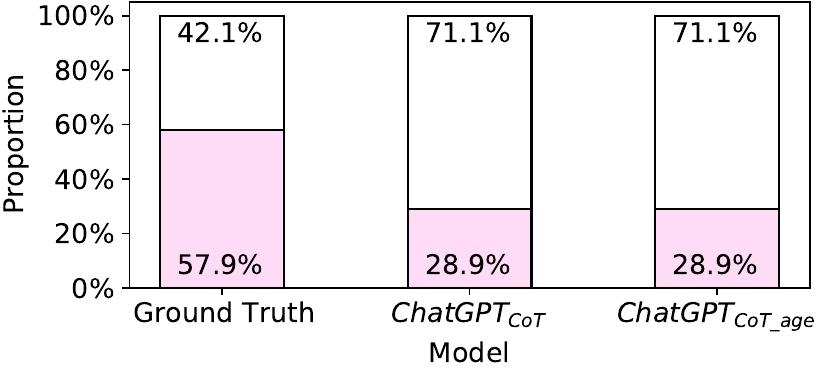}}
	\caption{Distribution of prediction results of ChatGPT$_{CoT}$ and ChatGPT$_{CoT\_age}$ towards age group (18 to 24), while purple and white denote low and high levels respectively.}\label{figure-distribution-prediction-age-1824}
\end{figure*} 

\begin{figure*}
	\centering
	\subfigure[Age (25 to 34) - O]{\includegraphics[width=0.32\textwidth]{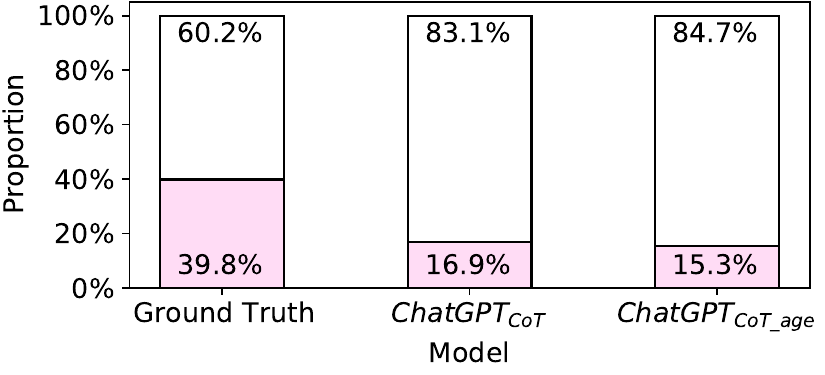}}
	\subfigure[Age (25 to 34) - C]{\includegraphics[width=0.32\textwidth]{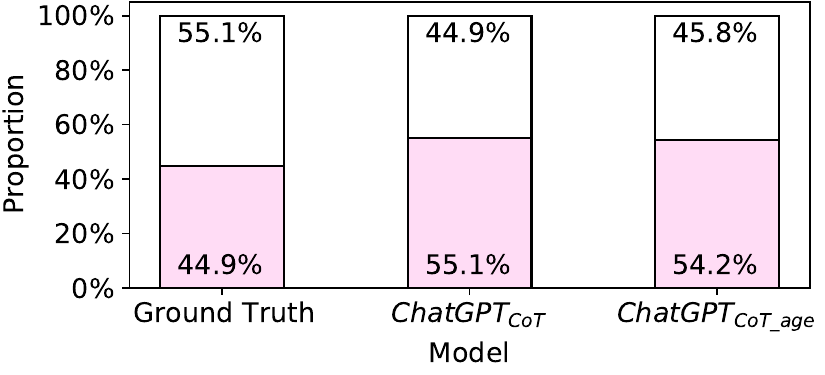}}
	\subfigure[Age (25 to 34) - E]{\includegraphics[width=0.32\textwidth]{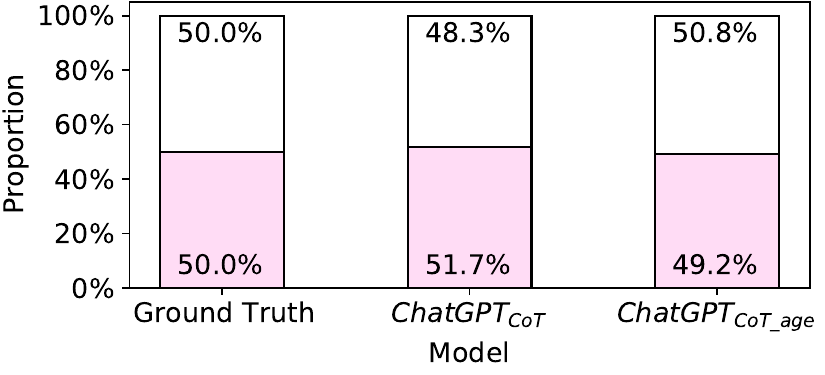}}
	\subfigure[Age (25 to 34) - A]{\includegraphics[width=0.32\textwidth]{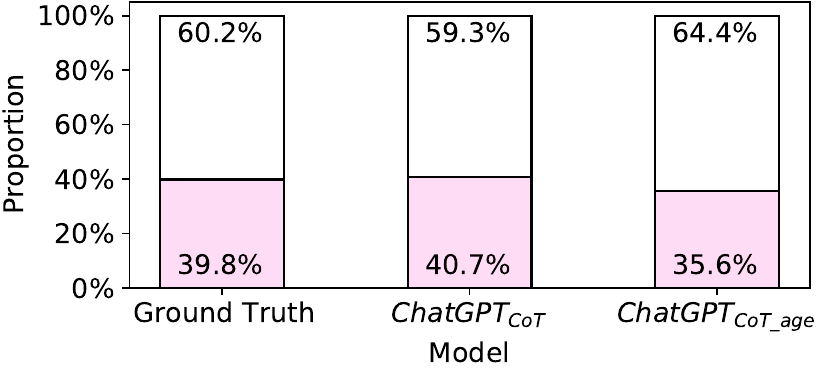}}
	\subfigure[Age (25 to 34) - N]{\includegraphics[width=0.32\textwidth]{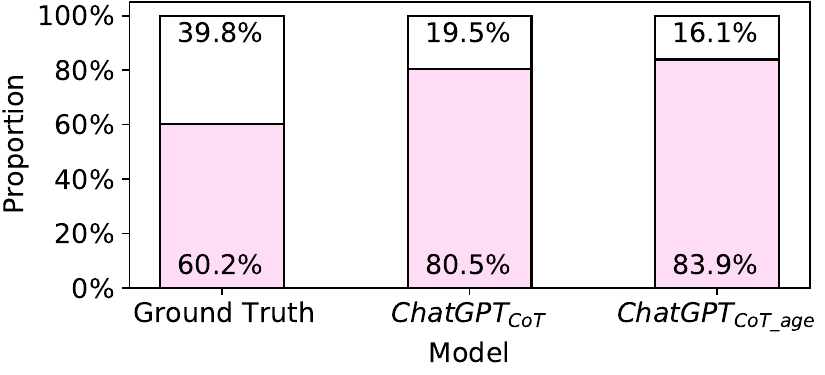}}
	\caption{Distribution of prediction results of ChatGPT$_{CoT}$ and ChatGPT$_{CoT\_age}$ towards age group (25 to 34), while purple and white denote low and high levels respectively.}\label{figure-distribution-prediction-age-2534}
\end{figure*}

\begin{figure*}
	\centering
	\subfigure[Age (35 to 49) - O]{\includegraphics[width=0.32\textwidth]{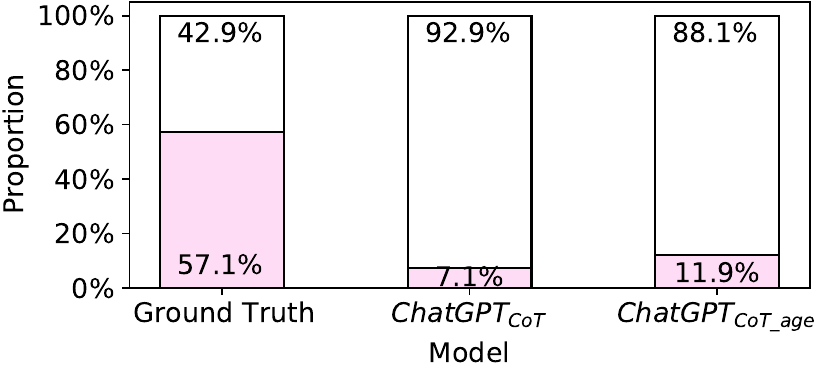}}
	\subfigure[Age (35 to 49) - C]{\includegraphics[width=0.32\textwidth]{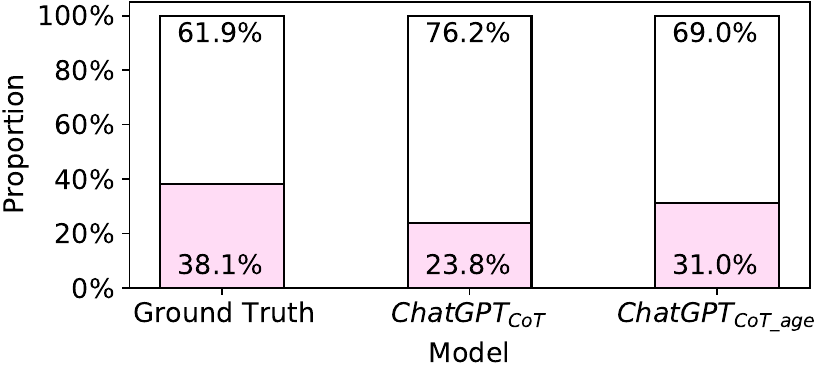}}
	\subfigure[Age (35 to 49) - E]{\includegraphics[width=0.32\textwidth]{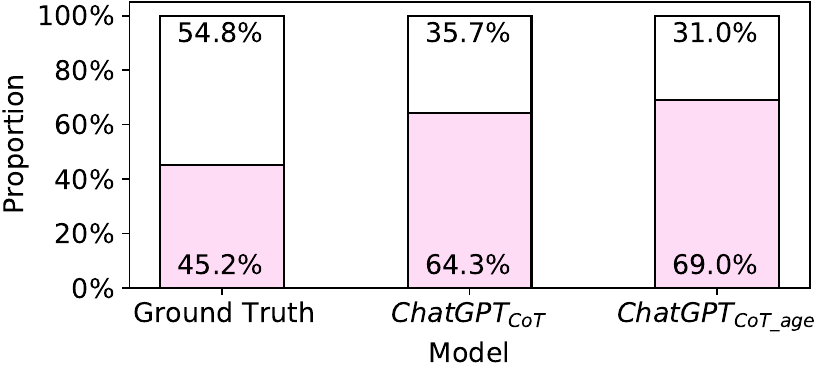}}
	\subfigure[Age (35 to 49) - A]{\includegraphics[width=0.32\textwidth]{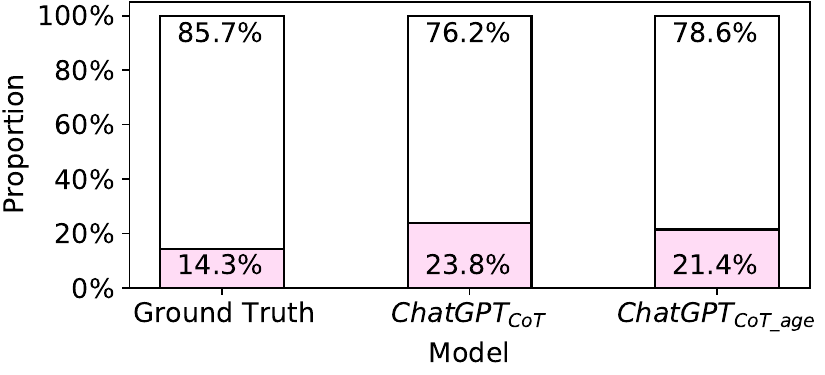}}
	\subfigure[Age (35 to 49) - N]{\includegraphics[width=0.32\textwidth]{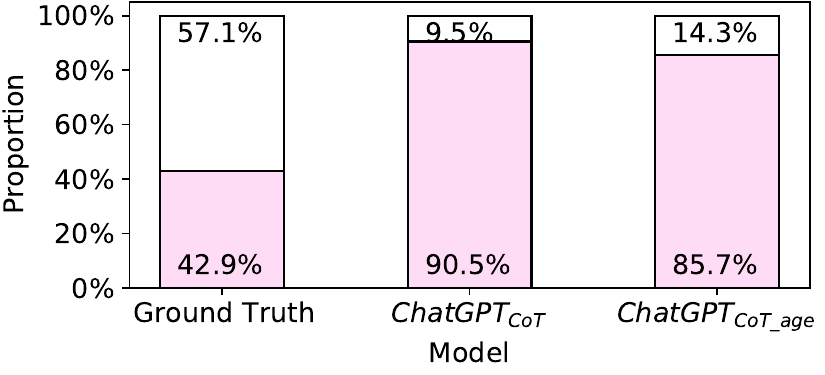}}
	\caption{Distribution of prediction results of ChatGPT$_{CoT}$ and ChatGPT$_{CoT\_age}$ towards age group (35 to 49), while purple and white denote low and high levels respectively.}\label{figure-distribution-prediction-age-3549}
\end{figure*}

\begin{figure*}
	\centering
	\subfigure[Age ($\geq$50) - O]{\includegraphics[width=0.32\textwidth]{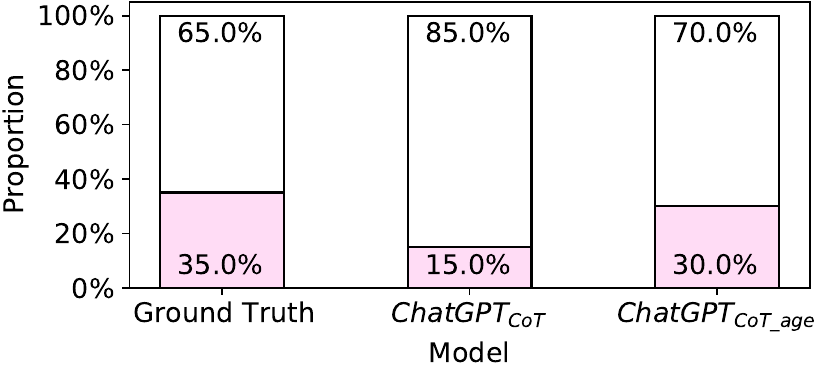}}
	\subfigure[Age ($\geq$50) - C]{\includegraphics[width=0.32\textwidth]{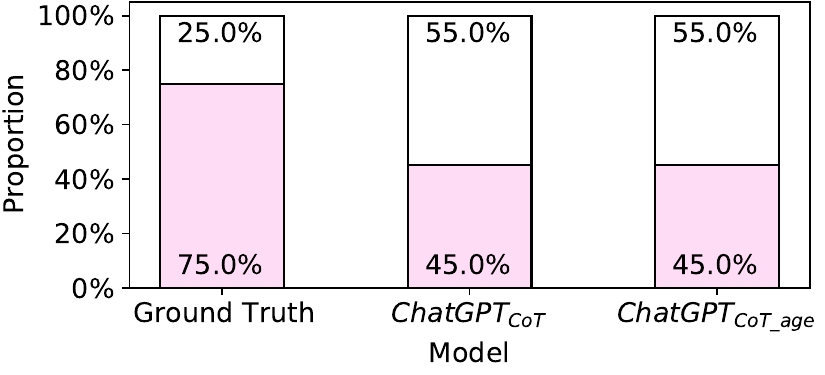}}
	\subfigure[Age ($\geq$50) - E]{\includegraphics[width=0.32\textwidth]{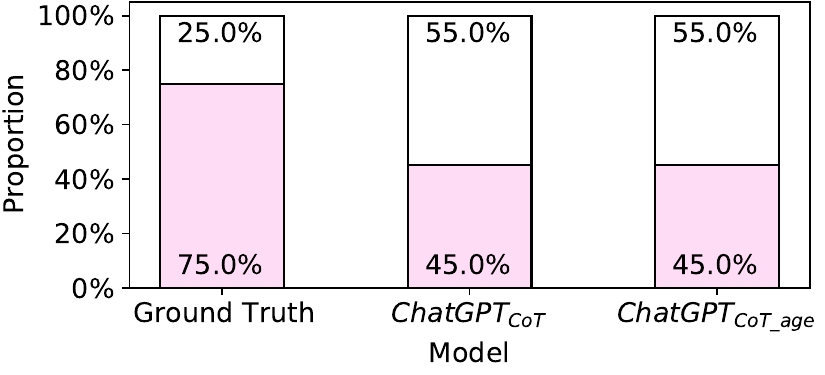}}
	\subfigure[Age ($\geq$50) - A]{\includegraphics[width=0.32\textwidth]{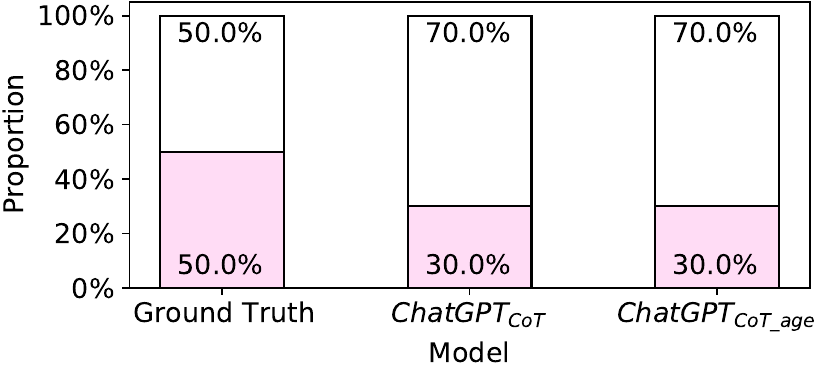}}
	\subfigure[Age ($\geq$50) - N]{\includegraphics[width=0.32\textwidth]{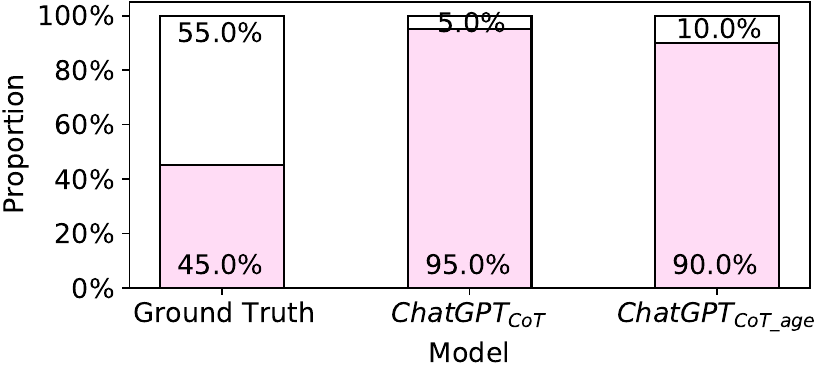}}
	\caption{Distribution of prediction results of ChatGPT$_{CoT}$ and ChatGPT$_{CoT\_age}$ towards age group ($\geq$50), while purple and white denote low and high levels respectively.}\label{figure-distribution-prediction-age-50}
\end{figure*}

\begin{figure*}
	\centering
	\subfigure[Gender (Woman)]{\label{figure-prediction-attribute-change-woman}\includegraphics[width=0.32\textwidth]{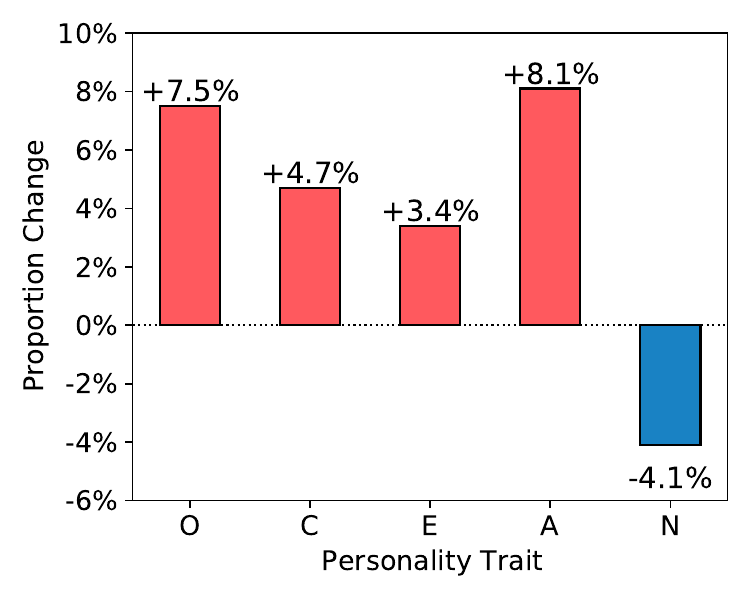}}
	\subfigure[Gender (Man)]{\includegraphics[width=0.32\textwidth]{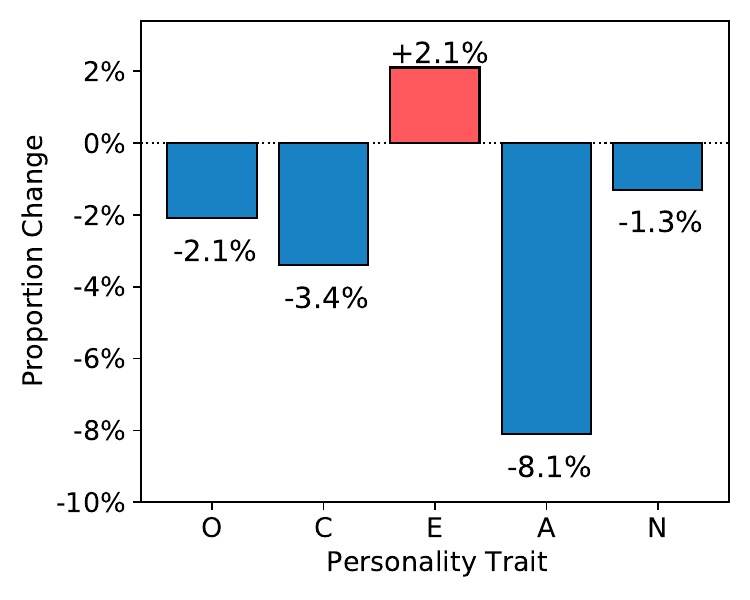}}
	\subfigure[Age (18 to 24)]{\includegraphics[width=0.32\textwidth]{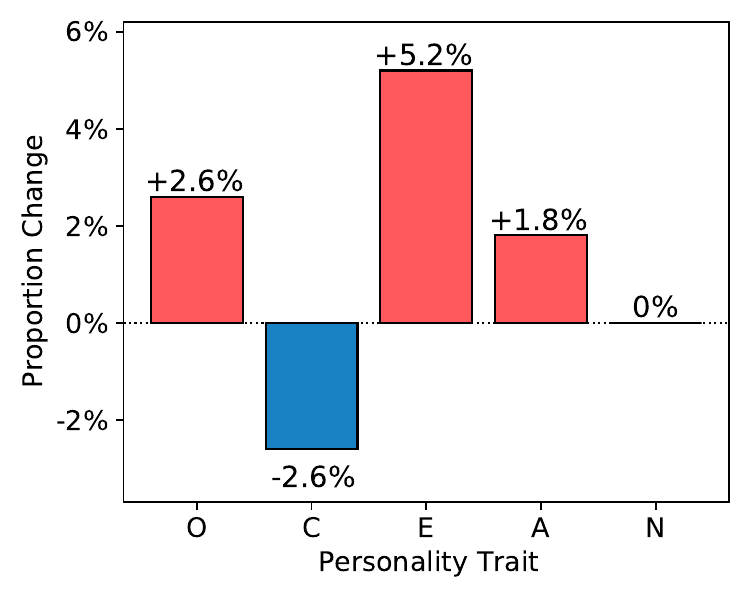}}

	\subfigure[Age (25 to 34)]{\includegraphics[width=0.32\textwidth]{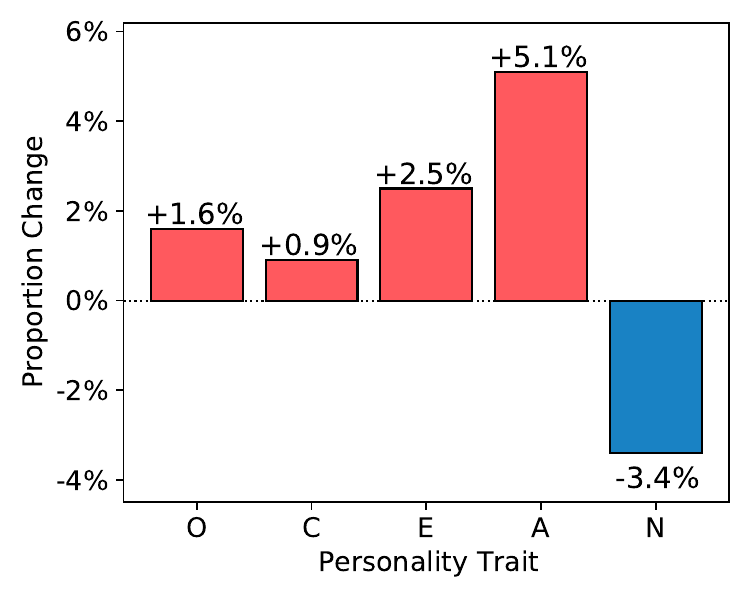}}
	\subfigure[Age (35 to 49)]{\includegraphics[width=0.32\textwidth]{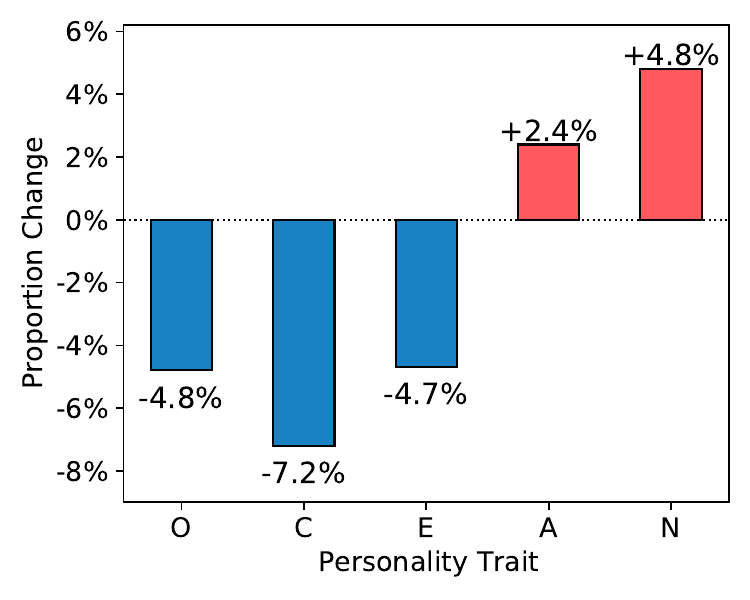}}
	\subfigure[Age ($\geq$50)]{\includegraphics[width=0.32\textwidth]{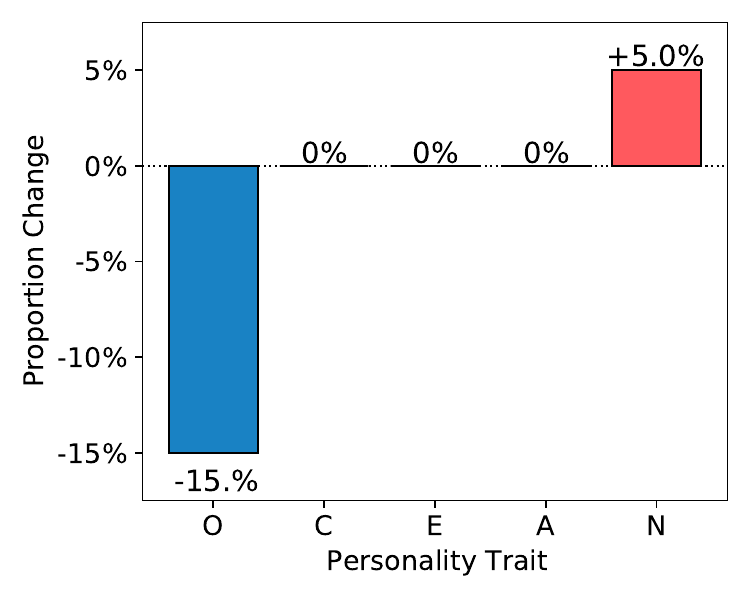}}
	\caption{The changes of the proportion of high level in the prediction results of ChatGPT$_{CoT\_gender}$/ ChatGPT$_{CoT\_age}$ relative to ChatGPT$_{CoT}$. Red indicates an increase in the proportion of high level, while blue indicates a decrease in the proportion of high level.}\label{figure-prediction-attribute-change}
\end{figure*} 

It is apparent from Figure~\ref{figure-attribute-results} that the incorporation of demographic attributes impairs the personality prediction ability of ChatGPT$_{CoT}$ to some extent, especially the integration of age information. For example, relative to ChatGPT$_{CoT}$, ChatGPT$_{CoT\_gender}$ and ChatGPT$_{CoT\_age}$ decrease their average accuracy from 55.5\% to 55.2\% and 54.0\% respectively. We speculate that this phenomenon may be due to ChatGPT's biases towards certain groups, which leads to unfair treatment of those groups. In order to better observe ChatGPT's biases on personality prediction task, we first obtain the prediction results of ChatGPT$_{CoT}$, ChatGPT$_{CoT\_gender}$, and ChatGPT$_{CoT\_age}$ towards different groups. We then visualize the proportion of low and high levels in those prediction results. Concretely, Figure~\ref{figure-distribution-prediction-gender-woman} and Figure~\ref{figure-distribution-prediction-gender-man} show the distribution of the prediction results of ChatGPT$_{CoT}$ and ChatGPT$_{CoT\_gender}$ towards woman and man groups respectively. In addition, Figure~\ref{figure-distribution-prediction-age-1824}, Figure~\ref{figure-distribution-prediction-age-2534}, Figure~\ref{figure-distribution-prediction-age-3549}, and Figure~\ref{figure-distribution-prediction-age-50} illustrate the distribution of the prediction results of ChatGPT$_{CoT}$ and ChatGPT$_{CoT\_age}$ towards different age groups. Take Figure~\ref{figure-distribution-prediction-gender-woman-O} as an example, the figure represents that among the 174 women in PAN dataset, 51\% of them have high O (i.e., ground truth). However, ChatGPT$_{CoT}$ classifies 74.8\% of the 174 women as high O, while ChatGPT$_{CoT\_gender}$ classifies 82.3\% of the 174 women as high O. In contrast, as shown in Figure~\ref{figure-distribution-prediction-gender-man-O}, among the 174 men in PAN dataset, 47.6\% of them have low O (i.e., ground truth). However, ChatGPT$_{CoT}$ classifies 29.9\% of the 174 men as low O, while ChatGPT$_{CoT\_gender}$ classifies 32.0\% of the 174 men as low O. In summary, after adding gender information, ChatGPT$_{CoT\_gender}$ classifies more women as high O and classifies more men as low O. This phenomenon suggests that ChatGPT considers women to be more likely to belong to high O when compared to men. In order to make a more intuitive comparison of the prediction results of ChatGPT$_{CoT}$, ChatGPT$_{CoT\_gender}$, and ChatGPT$_{CoT\_age}$ towards different groups, we further visualize the changes of the proportion of high level in the prediction results of ChatGPT$_{CoT\_gender}$/ ChatGPT$_{CoT\_age}$ relative to ChatGPT$_{CoT}$ (see Figure~\ref{figure-prediction-attribute-change}). For example, as displayed in Figure~\ref{figure-prediction-attribute-change-woman}, for 174 women in PAN dataset, the proportion of women with high A in the prediction results of ChatGPT$_{CoT\_gender}$ has increased by 8.1\% when compared to ChatGPT$_{CoT}$. Based on Figure~\ref{figure-prediction-attribute-change}, the biases of ChatGPT towards certain groups can be summarized as follows:

(1) Relative to the man group, the woman group is more likely to exhibit high levels of personality traits $O$, $C$, and $A$.

(2) The older an individual is, the greater the likelihood of her/his personality traits $O$ being low level.

However, these findings are not entirely consistent with existing research. For example, some studies suggest that the woman group is more likely to exhibit high levels of personality traits $A$ and $N$ compared to the man group, whereas gender differences in the other personality traits (i.e., $O$, $C$, and $E$) have been either inconsistent or of negligible magnitude \cite{lehmann2013age}. Possible reasons for this could be that, on the one hand, ChatGPT's biases are influenced by the biases of the annotators, which may not be representative. On the other hand, these findings are discovered based solely on the PAN dataset, limiting their generalization to some extent. Nevertheless, this phenomenon serves as a cautionary reminder for researchers to consider fairness when utilizing ChatGPT for personality prediction.

\subsection{ChatGPT's Personality Recognition Ability on Downstream Task ({\bf RQ3})}

\begin{table*}
\centering
\caption{Different promptings for sentiment classification task and stress prediction task}\label{table-down}
\begin{tabular}{lp{12.5cm}}
\toprule
Type               & Prompting                                                                                                                                                                                                                                                             \\ \midrule
\multicolumn{2}{l}{\textit{Sentiment Classification Task}} \\ \midrule   
Basic prompting           & Given this text, what is the sentiment conveyed? Is it positive or negative? Text: \{sentence\}                                                                                                                                                                          \\ \midrule
Modified basic prompting       & Given this text, what is the sentiment conveyed? Is it positive or negative? Note that, the person who generated the text has Low/High Openness, Low/High Conscientiousness, Low/High Extraversion, Low/High Agreeableness, and Low/High Neuroticism. Text: \{sentence\} \\ \midrule
\multicolumn{2}{l}{\textit{Stress Prediction Task}} \\ \midrule
Basic prompting     & Post: ``{[}Post{]}". Consider the emotions expressed from this post to answer the question: Is the poster likely to suffer from very severe stress? Only return Yes or No, then explain your reasoning step by step.                                                       \\ \midrule
Modified basic prompting & Post: ``{[}Post{]}". Consider the emotions expressed from this post to answer the question: Is the poster likely to suffer from very severe stress? Only return Yes or No, then explain your reasoning step by step. Note that, the poster has Low/High Openness, Low/High Conscientiousness, Low/High Extraversion, Low/High Agreeableness, and Low/High Neuroticism.                                                        \\ \bottomrule
\end{tabular}
\end{table*}

\begin{figure}
	\centering
	\subfigure[Sentiment classification task]{\label{figure-downstream-sentiment}\includegraphics[width=0.45\textwidth]{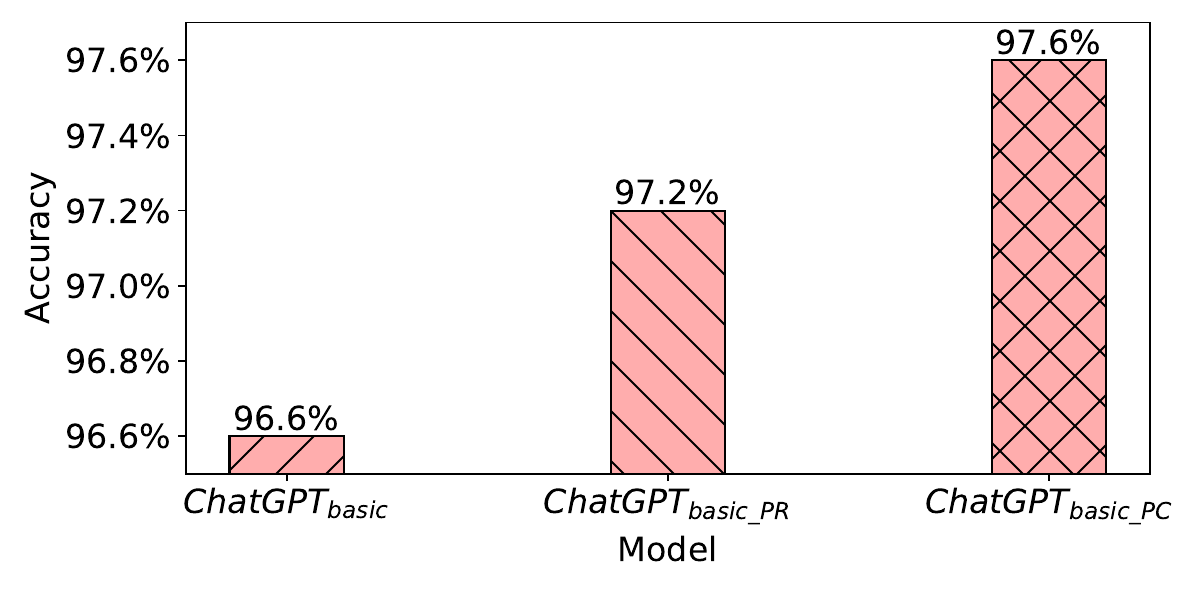}}
	\subfigure[Stress prediction task]{\label{figure-downstream-stress}\includegraphics[width=0.45\textwidth]{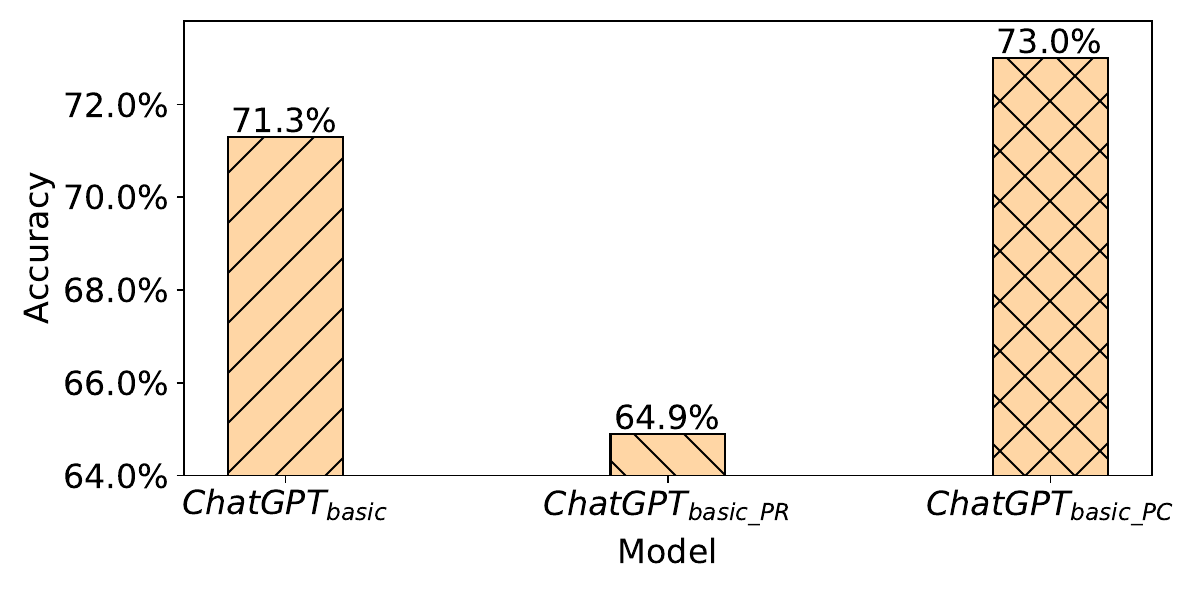}}
	\caption{The experimental results of sentiment classification task and stress prediction task.}\label{figure-downstream}
\end{figure}

We apply the personality data generated by ChatGPT to other downstream tasks for validating the effectiveness of ChatGPT's personality recognition ability. Concretely, we choose sentiment classification task and stress prediction task as the downstream tasks, because existing psychological research indicates that there is a correlation between Big-Five personality and sentiment expression \cite{kalat2011emotion} as well as stress vulnerability \cite{bunevicius2008symptoms}. For each task, to make a more comprehensive assessment of the impact of personality data generated by ChatGPT, we first adopt ChatGPT$_{CoT}$ and fine-tuned RoBERTa to generate the corresponding Big-Five personality based on given text respectively. We then use a basic prompting to elicit the task-related ability (i.e., sentiment classification ability and stress prediction ability) of ChatGPT. Finally, we modify the basic prompting by incorporating different Big-Five personalities and observe the task-related ability of ChatGPT with different modified basic promptings. 

To be specific, for sentiment classification task, we adopt a subset of Yelp-2 dataset \cite{zhang2015character} for conducting experiments. The reason for not utilizing the complete Yelp-2 dataset is to take into account the cost of using ChatGPT's API. Concretely, we randomly select 500 positive samples and 500 negative samples from the testing set of Yelp-2 dataset to construct the subset. While for stress prediction task, we choose Dreaddit dataset, which consists of 715 samples (369 positive samples and 346 negative samples) in its testing set. Specifically, considering that the texts in the PAN dataset, Yelp-2 dataset, and Stress dataset are all web posts, we use fine-tuned RoBERTa trained on PAN dataset to generate personality data. Besides, since both tasks are binary classification tasks, we adopt \textit{Accuarcy} (the higher the better) as the evaluation metric. In addition, the basic promptings used for sentiment classification task and stress prediction task are proposed by \cite{wang2023chatgpt} and \cite{yang2023evaluations}. Please refer to Table~\ref{table-down} for the detail of the unmodified/modified basic promptings. 

The experimental results are illustrated in Figure~\ref{figure-downstream}. Note that, ChatGPT$_{basic}$ represents ChatGPT with the basic prompting, while ChatGPT$_{basic\_PC}$ and ChatGPT$_{basic\_PR}$ denotes ChatGPT with the modified basic promptings, which incorporates the personality data generated by ChatGPT$_{CoT}$ and fine-tuned RoBERTa respectively. It can be observed that after incorporating the personality data predicted by ChatGPT$_{CoT}$, there is an improvement in ChatGPT's performance on both sentiment classification task and stress prediction task. For example, ChatGPT$_{basic\_PC}$ increases its classification accuracy from 96.6\% to 97.6\% on sentiment classification task when compared to ChatGPT$_{basic}$. While for stress prediction task, ChatGPT$_{basic\_PC}$ increases its classification accuracy from 71.3\% to 73.0\% when compared to ChatGPT$_{basic}$. This proves the effectiveness of the personality data generated by ChatGPT$_{CoT}$. With an understanding of individuals' Big-Five personalities, ChatGPT can analyze their sentiment expression and stress condition in a more personalized manner. Another interesting finding is that the personality data generated by fine-tuned RoBERTa can help improve the performance of ChatGPT in sentiment classification tasks, but it actually decreases ChatGPT's performance in stress prediction task. We believe that the possible reason for this is that fine-tuned RoBERTa trained on PAN dataset does not generalize well, which results in the poor performance of personality prediction on Dreaddit dataset. In contrast, ChatGPT relies solely on zero-shot CoT prompting to elicit its personality prediction ability and does not require training data, thus exhibiting stronger generalization performance on different datasets.

\section{Conclusion and Future Directions}\label{sec-conclusion}

In this work, we evaluate the personality recognition ability of ChatGPT with different prompting strategies, and compare its performance with RNN, fine-tuned RoBERTa, and corresponding SOTA model on two representative text-based personality identification datasets. With the elicitation of zero-shot CoT prompting, ChatGPT exhibits impressive personality recognition ability and has strong interpretability for its prediction results. In addition, we find that guiding ChatGPT to analyze text at a specified level helps improve its ability to predict personality, which proves the effectiveness of level-oriented prompting strategy. Moreover, we discover that ChatGPT exhibits unfairness to some sensitive demographic attributes, leading to unfair treatment of some specific groups when predicting personality. Besides, we apply the personality data inferred by ChatGPT in other downstream tasks and achieve performance improvement to some extent. This proves that ChatGPT's personality prediction ability is effective and has high generalization performance. 

As for future work, on the one hand, we would like to apply level-oriented prompting strategy to more NLP tasks for observing its effectiveness in mining text information. On the other hand, with the continuous emergence of various LLMs, we are interested in exploring the construction of domain-specific LLMs based on psychological data in order to enhance the personality recognition ability of LLMs.

\section*{Acknowledgment}
This work is funded by Science and Technology Commission of Shanghai Municipality, China (under project No. 21511100302), National Natural Science Foundation of China (under project No. 61907016), Natural Science Foundation of Shanghai (under project No. 22ZR1419000), the Research Project of Changning District Science and Technology Committee (under project No. CNKW2022Y37), and the Medical Master's and Doctoral Innovation Talent Base Project of Changning District (under project No. RCJD2022S07). In addition, it is also supported by The Research Project of Shanghai Science and Technology Commission (20dz2260300) and The Fundamental Research Funds for the Central Universities.

\subsection*{CRediT Authorship Contribution Statement}
{\bf Yu Ji:} Conceptualization, Methodology, Software, Validation, Formal analysis, Investigation, Data Curation, Writing-Original Draft, Writing-Review and Editing. {\bf Wen Wu:} Conceptualization, Methodology, Formal analysis, Investigation, Writing-Original Draft, Writing-Review and Editing, Supervision. {\bf Hong Zheng:} Writing-Review and Editing. {\bf Yi Hu:} Supervision, Writing-Review and Editing. {\bf Xi Chen:} Writing-Review and Editing. {\bf Liang He:} Supervision, Writing-Review and Editing.

\subsection*{Ethical Approval}
Not applicable.

\subsection*{Data Availability}
Data will be made available on request.

\subsection*{Declaration of Competing Interest}
The authors declare that they have no known competing financial interests or personal relationships that could have appeared to influence the work reported in this paper.

\bibliographystyle{unsrt}
\bibliography{cas-dc-template}

\end{document}